\documentclass[sigconf]{acmart}

\usepackage{booktabs} 

\usepackage{url, graphicx, subfigure}
\usepackage{amsmath, amsthm, amssymb}
\usepackage{algorithm,algorithmic,paralist}
\usepackage{bm}
\def \bx {\bm{x}}
\newcommand{\argmin}{\operatornamewithlimits{argmin}}
\usepackage{multirow}
\usepackage{threeparttable}

\copyrightyear{2018} 
\acmYear{2018} 
\setcopyright{acmcopyright}
\acmConference[KDD '18]{The 24th ACM SIGKDD International Conference on Knowledge Discovery \& Data Mining}{August 19--23, 2018}{London, United Kingdom}
\acmBooktitle{KDD '18: The 24th ACM SIGKDD International Conference on Knowledge Discovery \& Data Mining, August 19--23, 2018, London, United Kingdom}
\acmPrice{15.00}
\acmDOI{10.1145/3219819.3220026}
\acmISBN{978-1-4503-5552-0/18/08}

\begin{document}
\settopmatter{printacmref=false} 
\renewcommand\footnotetextcopyrightpermission[1]{} 
\pagestyle{plain} 

\title{Cost-Effective Training of Deep CNNs with Active Model Adaptation}

\author{Sheng-Jun Huang,\quad Jia-Wei Zhao,\quad Zhao-Yang Liu}
\affiliation{%
	\institution{College of Computer Science and Technology, Nanjing University of Aeronautics and Astronautics\\Collaborative Innovation Center of Novel Software Technology and Industrialization}
	\streetaddress{29 Jiangjun Road}
	\city{Nanjing 211106}
	\state{China}
	\postcode{211106}
}
\email{{huangsj, jiaweizhao,  zhaoyangliu}@nuaa.edu.cn}

\renewcommand{\shortauthors}{S.-J. Huang et al.}

\begin{abstract}
	Deep convolutional neural networks have achieved great success in various applications. 
	However, training an effective DNN model for a specific task is rather challenging because it requires a prior knowledge or experience to design the network architecture, repeated trial-and-error process to tune the parameters, and a large set of labeled data to train the model. In this paper, we propose to overcome these challenges by actively adapting a pre-trained model to a new task with less labeled examples. Specifically, the pre-trained model is iteratively fine tuned based on the most useful examples. The examples are actively selected based on a novel criterion, which jointly estimates the potential contribution of an instance on optimizing the feature representation as well as improving the classification model for the target task. On one hand, the pre-trained model brings plentiful information from its original task, avoiding redesign of the network architecture or training from scratch; and on the other hand, the labeling cost can be significantly reduced by active label querying. Experiments on multiple datasets and different pre-trained models demonstrate that the proposed approach can achieve cost-effective training of DNNs. 
\end{abstract}

\maketitle

\section{Introduction}
Deep neural networks have shown to be very effective in many fields. In addition to the impressive performance, it is also well known that training an effective deep model for a new task from scratch could be rather challenging \cite{8,11,long2017deep}. Firstly, a prior knowledge or rich experience is needed to design a suitable network architecture for a new task. Secondly, one should repeat the trial-and-error process to tune the hyper-parameters, which may significantly affect the learning performance. Lastly, a huge amount of labeled data is required for model training, leading to high annotation cost. These problems commonly occur, and cannot be overcome in most real applications, strongly limiting the application of deep learning to more tasks. It is thus important to have some effective strategies to train deep neural networks with lower cost on model designing, parameter optimization and data labeling.

To reduce the cost of designing network architecture and optimizing the model parameters, one straightforward idea is to exploit some pre-trained models instead of building the model from scratch. Transfer learning is an important method to ease the training of the target model by exploiting information from a different but related source domain \cite{5}. Under the deep neuron network framework, knowledge transfer from source domain to target domain has been implemented via distribution matching \cite{2}, feature sharing \cite{4} or components transformation \cite{pan2011domain}. However, these methods typically revise the architecture of the pre-trained networks, making the retraining still challenging with too many parameters. A better choice is to perform model adaptation with fixed architecture of pre-trained models. For example, in computer vision field, we are lucky to have a large dataset ImageNet \cite{deng2009imagenet} with all images manually labeled. Some well-known models such as AlexNet \cite{18}, VGG \cite{19} and ResNet \cite{20} have been pre-trained on ImageNet with descent performance. These models are good at learning effective representations for visual objects, and are public available for us to utilize. The simplest way to utilize the pre-trained model is to directly employ the entire network except for the output layer as a feature extractor. However, such a simple strategy is less practical because the feature extracted will be less effective when the target task is not very similar to the original one \cite{1}. One alternative approach is to retain the architecture of the pre-trained model, and then retrain the model via fine-tuning its weights. Such methods can partially reduce the training cost, but still require a relatively large dataset to optimize the network weights \cite{6}.

Active learning is a primary approach for reducing the labeling cost \cite{7,HJZ14}. It iteratively queries the labels for the most useful instances, and tries to train an effective model with less queries. Various criteria have been proposed for active selection in traditional shallow models \cite{28,chattopadhyay2013joint,CL2015b}. For example, informativeness and representativeness have been demonstrated to be good choice for estimating the potential contribution of an instance on improving the classification \cite{HJZ14,wang2015querying,HZ13}. However, the active learning strategies designed for traditional shallow models have been validated to be not effective for deep models, because of the conflict between the batch sampling redundancy and local optimization \cite{10}. There are several studies focus on active learning for deep neural networks \cite{10,11,12}. However, they usually do not consider the pre-trained model, and thus may lead to waste of annotation cost by querying information already contained in pre-trained models.

In this paper, we propose to perform active model adaptation for cost-effective training of deep convolutional neural networks. On one hand, the training cost is reduced by fine tuning the pre-trained models with frozen layers; and on the other hand, the labeling cost is reduced by actively querying the labels for the most useful instances. Specifically, a novel criterion for active selection is proposed to simultaneously consider and dynamically adjust the distinctiveness ( peculiarity  endemism) and uncertainty of an instance. Thus the selected instances are expected to be most useful for both the classifier training and representation learning.

We perform experiments on multiple datasets and different pre-trained models. The results demonstrate that the proposed approach can effectively train a deep convolutional neural network model with significantly lower cost both on the training process and the label annotation. Specifically, our algorithm can achieve comparable performance by using only 5\% of the training data compared to passive training from scratch.

The main contributions of this work are summarized as follows.
\begin{itemize}
	\item A general framework of active model adaptation for deep convolutional neural networks. It can be applied to different pre-trained models and incorporated with various active selection strategies. 
	\item A novel criterion \emph{distinctiveness} is proposed to measure the potential contribution of an instance on improving the feature representation of the network.
	\item An algorithm can actively select instances based on dynamic trade-off between \emph{distinctiveness} and \emph{uncertainty}, and can also improve the deep model to achieve better feature representation and label prediction.
	\item Extensive experiments validate the effectiveness of the proposed approach.
\end{itemize}

The rest of this paper is organized as follows. In Section 2, we review the previous work. In Section 3, the propose approach is introduced. Section 4 presents the experiments, followed by the conclusion in Section 5.

\section{Related Work}
Transfer learning tries to improve the performance in a target domain by transferring information from a related but different source domain \cite{5,27}. It has been extensively studied by transferring knowledge from source domain to target domain at instance, feature or model levels \cite{5,gong2013connecting,zhang2013domain,duan2012domain,sugiyama2008direct}. Among which, model level transferring is more related to our work \cite{26,pan2011domain}. Generally speaking, these methods try to train a new model for the target domain by directly reusing or modifying the model that has been well trained in the source domain. 

Recently, there are increasing interests on transfer learning for deep neural networks \cite{long2017deep,ganin2015unsupervised}. For example, authors of \cite{2} propose a new network structure which adds an adaptation layer to minimize both domain confusion loss and classification loss simultaneously. In \cite{4}, domain adaptation is considered in all task-specific layers for matching different domain distributions effectively. Inspired by WGAN\cite{30}, in \cite{29},  the domain discrepancy based on Wasserstein distance is minimized in adversarial learning process to optimize the representation.   

These methods can partially ease the training of deep models, but they do not consider active learning during the transfer process, and thus still need a relatively large set of labeled examples to train an effective model for the target domain. Moreover, they usually need to revise the architecture of the network, and thus cannot overcome the challenge on expensive network design and optimization.

Active learning reduces the labeled training examples by selecting the most valuable instances to query their labels \cite{7}. During the past decades, many criteria have been proposed for active selection of instances \cite{HJZ14,28,7,wang2015convex}. Recently, there are some studies applying active learning strategies into deep learning to reduce the training data \cite{8,10,11,12}. In \cite{8}, a tractable approximation method is proposed to estimate the prediction variance of a Bayesian CNNs \cite{9}, based on which, the examples with high variance are selected for label querying. The authors of \cite{10} transform active learning into a core-set selection problem under the deep learning setting, and try to select instances to make the model trained on the selected subset competitive for other examples. The method proposed in \cite{11} tries to query labels for uncertain instances, while at the same time assigns pseudo labels for the examples with high prediction confidence. In \cite{12},  the similarity between examples at the last layer of a fully convolutional network is combined with uncertainty for active selection.

There are some studies incorporating active queries for transfer learning with traditional shallow models. For example, the method in \cite{WHS14} combines active learning and transfer learning into a Gaussian Process based approach, and sequentially selects query points from the target domain based on the predictive covariance. Kale and Liu \shortcite{KL13} propose a principled framework to combine the agnostic active learning with transfer learning, and utilize labeled data from source domain to improve the performance in the target domain. Kale et al. \shortcite{KGRHL15} present a hierarchical framework to exploit cluster structure between different domains, and try to impute labels for unlabeled data and select active queries in the target domain based on the structure. Huang and Chen \shortcite{HC16} propose to actively query labels from source domain to help the learning task of the target domain. These methods are generally designed for traditional shallow models, and neglect the special challenges of training deep neural networks. 

There is one study trying to actively fine tune a pre-trained deep neural network for biomedical image analysis \cite{13}. The authors propose a new criterion to estimate the diversity among different patches extracted from the same image, and expect the image with more diverse patches is more useful for updating the model. However, this method is specially designed for biomedical image analysis, and can only handle binary classification problems, which strongly restricts its application.

\section{The Proposed Approach}
We denote by $U=\{\bm{x}_j\}_{j=1}^{n_u}$ the unlabeled dataset with $n_u$ instances, where $\bm x_{i}$ is the $i$-th instance. In this paper, we perform batch-mode active learning. At each iteration, a small batch of instances $Q=\{\bm{x}_q\}_{q=1}^b$ with size $b$ will be selected from $U$ to query their labels. We also denote by $\mathcal{M}^0$ the pre-trained model, and $\mathcal{M}^t$ the model at the $t$-th iteration. In the following subsections, we will first introduce the batch-mode active learning framework for deep CNN model adaptation, then propose the criterion for selecting instances, and at last summarize the main steps of the algorithm.

\subsection{The framework}
\begin{figure}[!t]
	\begin{center}
		\begin{minipage}{0.99\linewidth}
			\includegraphics[width=\textwidth]{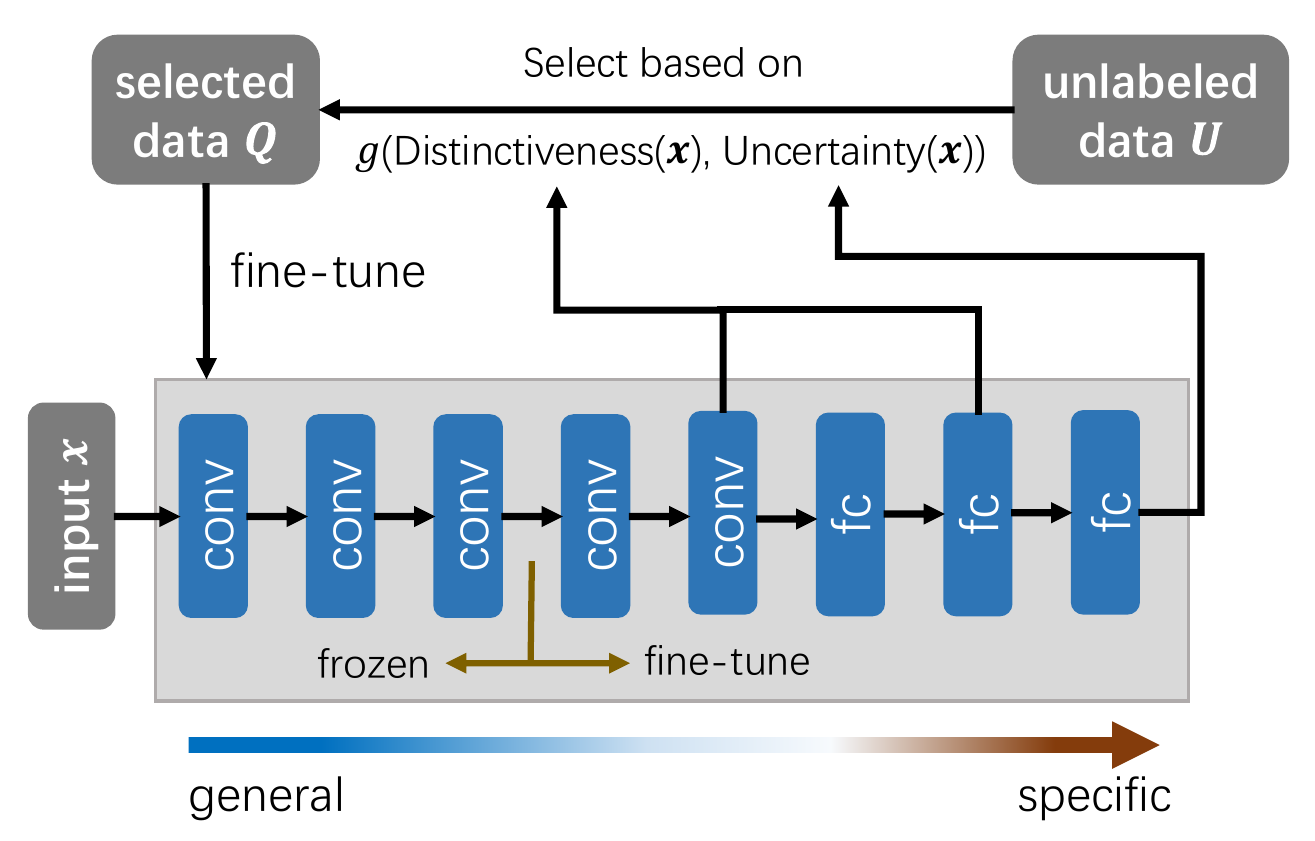}\\
		\end{minipage}
		\caption{The framework of active model adaptation for deep neural networks.}\label{fig:frame}
	\end{center}
\end{figure}

We focus on batch-mode active learning, i.e., query labels for a small batch of instances selected from the unlabeled set. Figure \ref{fig:frame} presents the framework of active model adaptation for deep convolutional neural networks. Here we take AlexNet \cite{18} as an example, which is a well-known model pre-trained on ImageNet ILSVRC dataset. Typically, the deep neural networks learn an effective representation from general to specific. The first few layers mainly capture universal features such as curves and edges, and then the later layers generate more task-specific features. The general features work for different tasks, while the specific features focus on describe the unique properties of a task. When adapting a pre-trained model from the source task to the target task, it is reasonable to retain the layers for general features, while update the specific layers to fit the requirement of the target task. 

For each instance in the unlabeled set, the original features are input into the current neural network. Then based on the representations in different layers and the final prediction outputs, distinctiveness and uncertainty of the input instance are estimated respectively. The distinctiveness measures the ability of an instance on capturing the particular property of the target task that differs from the source task; while the uncertainty measures the ability of an instance on improving the classification model. In other words, distinctive instances are responsible to improve the layers of learning the representation; while uncertain instances are responsible to improve the last layer of classification. The two criteria are jointly considered to select a small batch of most informative instances. After querying their labels, the selected instances are utilized to fine tune the neural network, where the early layers are frozen to retain the information from source task.

\begin{figure*}[!t]
	\begin{center}
		\begin{minipage}{0.99\linewidth}
			\includegraphics[width=\textwidth]{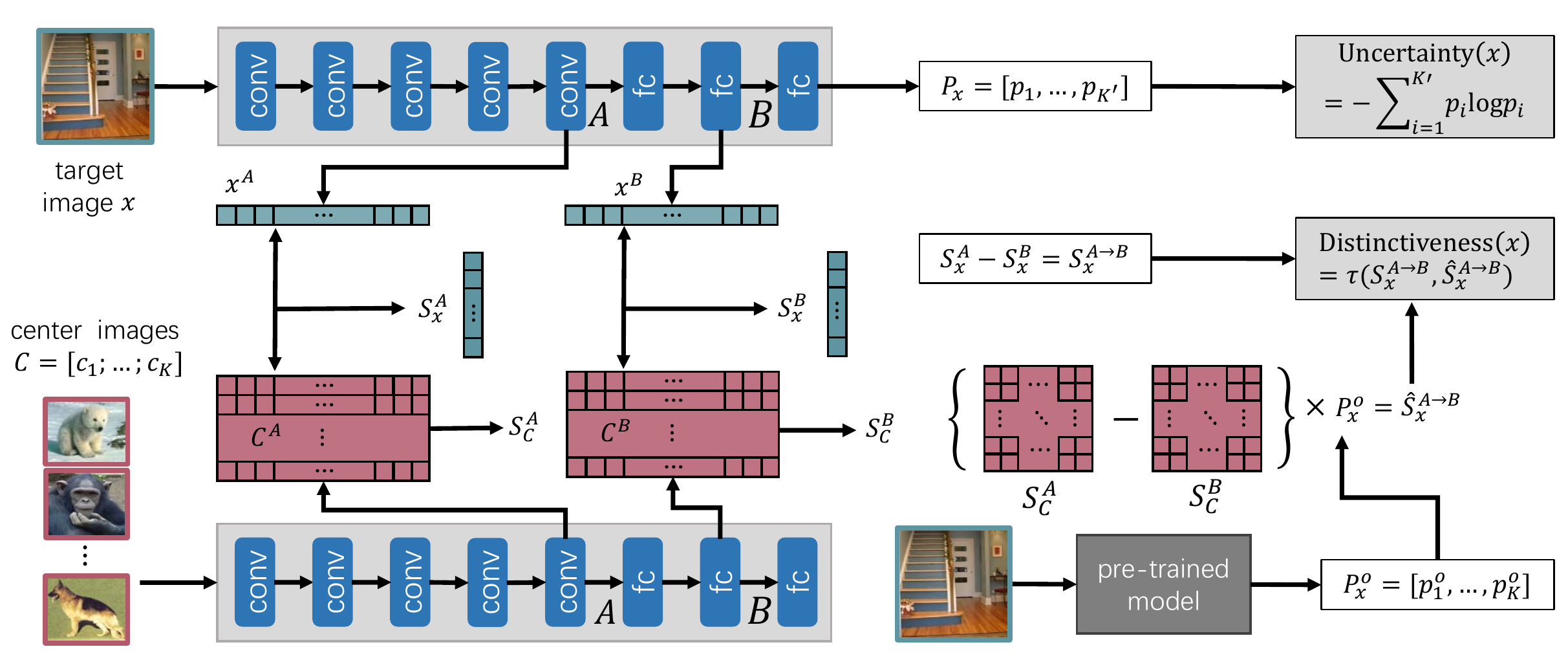}\\
		\end{minipage}
		\caption{The criterion for active selection.}\label{fig:criterion}
	\end{center}
\end{figure*}

\subsection{The active selection criterion}
In this subsection, we will introduce the two criteria \emph{distinctiveness} and \emph{uncertainty} respectively. In a deep convolutional neural network, the early layers produce general features, while later layers generate more specific features, and the last layer usually corresponds to a classifier. To adapt a deep network from a source task to a target task with fixed architecture, the key problem is to update the network weights. When actively selecting the instances to help the model adaptation, the model after adaptation is expected to on one hand, the adapted model is capable of learning an effective representation for the target task; and on the other hand, the classifier can achieve high accuracy. We thus propose two criteria to estimate the usefulness of an instance on these two aspects, which are named \emph{distinctiveness} and \emph{uncertainty} respectively. The workflow of calculating these two criteria is summarized in Figure \ref{fig:criterion}

\subsubsection{Distinctiveness}
We propose a novel criterion \emph{distinctiveness} to measure the ability of an instance on improving the representation quality of the neural network for the target task. The pre-trained model is optimized for the source task. If we want to optimize it to fit the target task, then some instances which can capture the unique property of the target task should be used to fine tune the model. We call such capability as distinctiveness because it distinguishes the target task from the source target. To estimate the distinctiveness of an instance, the basic idea is to firstly exploit the pattern of the pre-trained model on feature transformation from early to later layers. If an instance in the target task has a transformation pattern that is significantly different from that of the source task, then this instance is expected to be more distinctive. Following we introduce the detailed steps of calculating the distinctiveness of an instance.

First of all, assume that there are in all $K$ classes in the source task, then one representative instance is selected for each class, leading to the representative instance set $C=\{\bm c_1, \bm c_2, \cdots, \bm c_K\}$. We denote by $\bm z$ the final representation of an instance in the source task. Then for each class $k$, the mean of all instances belong this class can be calculated as
\begin{equation}
\bar{\bm{z}}_k=\frac{1}{|\Omega_k|}\sum_{\bm z\in \Omega_k}\bm{z},
\end{equation}
where $\Omega_k$ is the set of instances belong to the $k$-th class, and $|\cdot|$ calculates the set size. Then $\bm c_k$ is selected as 
\begin{equation}
\bm c_k=\argmin_{\bm z\in \Omega_k}\|\bm z-\bar{\bm z}_k\|^2.
\end{equation}
The feature transformation pattern on these representative centers is used to describe how the network learns the features from early layers to later layers in the source task. For convenience of presentation, we specify one earlier layer as $A$, and a latter layer as $B$, and then denote by $S_{\bm c}^{A\rightarrow B}$ the transformation pattern of the center $\bm c$. These transform patterns can be considered to be optimal for the source task, but less optimal for the target task. Given an instance $\bx$ in the target domain, a similar feature transform pattern $S_{\bm x}^{A\rightarrow B}$ can be obtained. In addition, we can further get an approximated pattern $\hat{S}_{\bm x}^{A\rightarrow B}$ by weighted combination of the representative source patterns. In fact, $\hat{S}_{\bm x}^{A\rightarrow B}$ is estimating how the network will transform the features of $\bx$ if it is an instance from source task, while $S_{\bm x}^{A\rightarrow B}$ is the observed pattern of $\bx$. So the difference between these two patterns reflects how distinctive $\bx$ is from the source task. Follows we discuss how to calculate feature transform pattern from layer $A$ to layer $B$.

Given a source center $\bm c_k$, its outputs at layer $A$ and layer $B$ are denoted by $\bm c_k^A$ and $\bm c_k^B$, respectively. Similarly, we have $\bx^A$ and $\bx^B$ for an target instance $\bx$. Then we define:
\begin{align}\label{eqn:SCA}
S_C^A=&\left[S_{\bm c_1}^A, S_{\bm c_2}^A\cdots, S_{\bm c_K}^A
\right]\nonumber\\
=&
\left[
\begin{array}{cccc}
\|\bm c_1^A-\bm c_1^A\|^2,& \|\bm c_2^A-\bm c_1^A\|^2& \cdots &\|\bm c_K^A-\bm c_1^A\|^2\\
\|\bm c_1^A-\bm c_2^A\|^2,& \|\bm c_2^A-\bm c_2^A\|^2& \cdots &\|\bm c_K^A-\bm c_2^A\|^2\\
\cdots& \cdots & \cdots & \cdots\\
\|\bm c_1^A-\bm c_K^A\|^2,& \|\bm c_2^A-\bm c_K^A\|^2& \cdots &\|\bm c_K^A-\bm c_K^A\|^2\\
\end{array}
\right],
\end{align}
and
\begin{equation}\label{eqn:SxA}
S_{\bx}^A=\left[
\begin{array}{c}
\|\bx-\bm c_1^A\|^2\\
\|\bx-\bm c_2^A\|^2\\
\cdots\\
\|\bx-\bm c_K^A\|^2\\
\end{array}
\right].
\end{equation}
Similarly, we have $S_C^B$ and $S_{\bx}^B$ corresponding to the layer $B$. In fact, the centers $\{\bm c_k\}_{k=1}^K$ are taken as landmarks, while $S_{\bx}^A$ and $S_{\bx}^B$ are capturing the relative representation at layers $A$ and $B$ based on the landmarks. Then, the feature transformation pattern from layer $A$ to layer $B$ can be simply obtained with the subtraction between the relative representation of two layers:
\begin{equation}
S_{\bm x}^{A\rightarrow B}=S_{\bx}^A -S_{\bx}^B.
\end{equation}
Next, we try to explore how would be the transformation pattern from layer $A$ to layer $B$ if $\bx$ was a instance from the source task instead of target task. Firstly, we can have the transformation pattern of each representative center $\bm c_k$ in the source task by taking all the other centers as landmarks, i.e.,
\begin{equation}
S_{\bm c_k}^{A\rightarrow B}=S_{\bm c_k}^A -S_{\bm c_k}^B.
\end{equation}
Then we try to approximate the transformation pattern of $\bx$ by a weighted linear combination of $\{S_{\bm c_k}^{A\rightarrow B}\}_{k=1}^K$. Formally,
\begin{equation}\label{eqn:Shx}
\hat{S}_{\bm x}^{A\rightarrow B}=\sum_{k=1}^K \alpha_k(\bx)\cdot S_{\bm c_k}^{A\rightarrow B},
\end{equation}
where $\alpha_k(\bx)$ is the weight corresponding to the $k$-th center. Here we define it as the probability of $\bx$ belongs to the $k$-th class based on the prediction of the original pre-trained model, i.e.,
\begin{equation}\label{eqn:alpha}
\alpha_k(\bx)=p(\mathcal{M}^0(\bx)=k),
\end{equation}
where $\mathcal{M}^0$ denotes the predicted class of $\bx$ by the pre-trained model $\mathcal{M}^0$. We will discuss and compare other possible implementations of $\alpha_k(\bx)$ in the experiments.

By now we have got $S_{\bm x}^{A\rightarrow B}$ as the observed feature transformation pattern of $\bx$, and $\hat{S}_{\bm x}^{A\rightarrow B}$ as the approximated transformation pattern by assuming it as an instance from the source task. The difference between these two patterns reflects the potential contribution to the model adaptation from source task to target task with regard to the representation learning, and is taken to estimate the distinctiveness of $\bx$. Noticing that the two patterns actually are vectors, and there could be various ways to estimate the difference between them. In our case here, because the relative rank correlation is more important than the exact value comparison, we employ the Kendall's tau coefficient \cite{knight1966computer} to estimate the difference, and finally have the definition of the distinctiveness as
\begin{equation}
{Distinctiveness}(\bx)= \frac{1-\tau(S_{\bm x}^{A\rightarrow B},\ \hat{S}_{\bm x}^{A\rightarrow B})}{2}
\end{equation}
where $\tau(S_{\bm x}^{A\rightarrow B},\ \hat{S}_{\bm x}^{A\rightarrow B})$ is the Kendall's tau coefficient between $S_{\bm x}^{A\rightarrow B}$ and $\hat{S}_{\bm x}^{A\rightarrow B}$.

\subsubsection{Uncertainty}

Uncertainty is a commonly used criterion in active learning to estimate how uncertain the prediction of the current model is for a given instance.  Assume that there are $K'$ classes in the target task. The uncertainty of $\bx$ is defined as:
\begin{equation}\label{eqn:uncertainty}
Uncertainty(\bx)=-\sum_{k'=1}^{K'} p(\mathcal{M}(\bx)=k')\cdot(1-p(\mathcal{M}(\bx)=k')),
\end{equation}
where $\mathcal{M}$ is the current model, and $p(\mathcal{M}(\bx)=k')$ is the probability of $\bx$ belongs to class $k'$ based on the prediction of $\mathcal{M}$. In addition to the entropy in Eq. \ref{eqn:uncertainty}, there could be other definitions of the uncertainty criterion, such as the margin to decision boundary or prediction confidence.

\subsubsection{Dynamic trade-off}
As discussed before, the \emph{distinctiveness} measures the contribution of an instance on improving the network for better representation learning, while the \emph{uncertainty}  measures the contribution on improving the classifier corresponding to the last layer of the neural network. To select the most useful instances for better adaptation of the pre-trained network to the target task, we should simultaneously consider the distinctiveness and uncertainty, such that the network can learn better representations as well as stronger classifier for the target task.

One key superiority of deep neural networks to traditional shallow models is that they can learn effective feature representations automatically. During the adaptation of the network, we believe the \emph{distinctiveness} and \emph{uncertainty} play different roles at different stages. At the beginning, the neural network is pre-trained for the source task, and the representation may be less effective for the target task. It is thus urgent to improve the representation at this stage. Moreover, based on the less effective features, it is meaningless to optimize the classifier of the last layer. So we should query more distinctive instances to improve the model towards better feature learning. After more and more queries, the representation is expected to be well adapted to the target task, and thus more uncertain instances are expected to improve the model towards better classification. Based on this motivation, we emphasize more on the \emph{distinctiveness} during the early stages of model adaptation, and gradually increase the attention on \emph{uncertainty}. Formally, we employ a trade-off parameter to dynamically balance the two criteria as the iterations progress:
\begin{equation}\label{eqn:score}
score(\bx)=(1-\lambda\cdot t)\cdot distinctiveness(\bx)+\lambda\cdot t \cdot uncertainty(\bx),
\end{equation}
where $t$ is the iteration number.

Finally, the most useful instances with large values of $score(\cdot)$ will be selected to query their labels, and further used to fine tune the network.

\subsection{The ADMA algorithm}
The main steps for the proposed ADMA (Active Deep Model Adaptation) algorithm are summarized in Algorithm \ref{al:adma}. Firstly, in the source task, one representative center instance is determined for each class. Then the feature transformation patterns from layer $A$ to layer $B$ are calculated for all the centers. After that, the algorithm performs batch-mode active learning to query labels and fine tunes the network iteratively. In the $t$-th iteration, for each unlabeled instance $\bx\in U$, its feature transformation pattern as well as the approximated pattern are calculated, based on which, the \emph{distinctiveness} is further obtained. Then, by combining the \emph{distinctiveness} and \emph{uncertainty}, the final \emph{score} to evaluate the potential contribution of $\bx$ is calculated. The unlabeled instances are ranked with regard to this score in descending order, and the top batch $Q$ is selected to query the labels. After that, the queried instances are used to fine tune the network from $\mathcal{M}^{t-1}$ to $\mathcal{M}^t$. This process is repeated until the the query budget is out or the network achieves a specified performance.

\begin{algorithm}[!tb]
	\caption{The ADMA algorithm}\label{al:adma}
	\label{algorithm}
	\begin{algorithmic}[1]
		\STATE \textbf{Input:}
		\STATE \quad $U$: the unlabeled set of $n_u$ instances
		\STATE \quad $\mathcal{M}^0$: the pre-trained model
		\STATE \quad $A$: the index of the initial layer for feature transformation
		\STATE \quad $B$: the index of the end layer for feature transformation
		\STATE \quad $Z$: the dataset in the source task used for training $\mathcal{M}^0$
		\STATE \textbf{Initialization:}
		\begin{ALC@g}
			\STATE Find the centers $C=\{\bm c_1, \cdots,\bm c_K\}$, one for each class $k$ in $Z$;
			\STATE	Obtain $\bm c^A_k$ and $\bm c^B_k$: the outputs of $\bm c_k$ at layers $A$ and $B$;
			\STATE Calculate $S_C^A$ and $S_C^B$ according to Eq. \ref{eqn:SCA};
			\STATE Calculate $S_C^{A\rightarrow B}=S_C^A-S_C^B$;
		\end{ALC@g}
		\STATE \textbf{Repeat:}
		\begin{ALC@g}
			\STATE \textbf{For} each instance $\bx\in U$		
			\begin{ALC@g}
				\STATE	Obtain $\bx^A_k$ and $\bx^B_k$: the outputs of $\bx$ at layers $A$ and $B$;
				\STATE Calculate $S_{\bx}^A$ and $S_{\bx}^B$ according to Eq. \ref{eqn:SxA};
				\STATE Calculate $S_{\bx}^{A\rightarrow B}=S_{\bx}^A-S_{\bx}^B$;
				\STATE Calculate $\hat{S}_{\bx}^{A\rightarrow B}$ by weighted summarization of $S_C^{A\rightarrow B}$;
				\STATE Calculate ${Distinctiveness}(\bx)= \frac{1-\tau(S_{\bm x}^{A\rightarrow B},\ \hat{S}_{\bm x}^{A\rightarrow B})}{2}$;
				\STATE Calculate $Uncertainty(\bx)$ according to Eq. \ref{eqn:uncertainty};
				\STATE Calculate the criterion $score(\bx)$ according to Eq. \ref{eqn:score};
			\end{ALC@g}
			\STATE \textbf{End For}
			\STATE Select a batch of instances $Q$ from $U$ with largest $score$;
			\STATE Query the labels for $Q$, and remove $Q$ from $U$;
			\STATE Fine-tune the model $\mathcal{M}^{t-1}$ with the queried data to get $\mathcal{M}^t$.
		\end{ALC@g}
		\STATE \textbf{Until} query budget or expected performance reached.
	\end{algorithmic}
\end{algorithm}

\begin{figure*}[!t]
	\begin{center}
		\begin{minipage}{0.32\linewidth}
			\includegraphics[width=\textwidth]{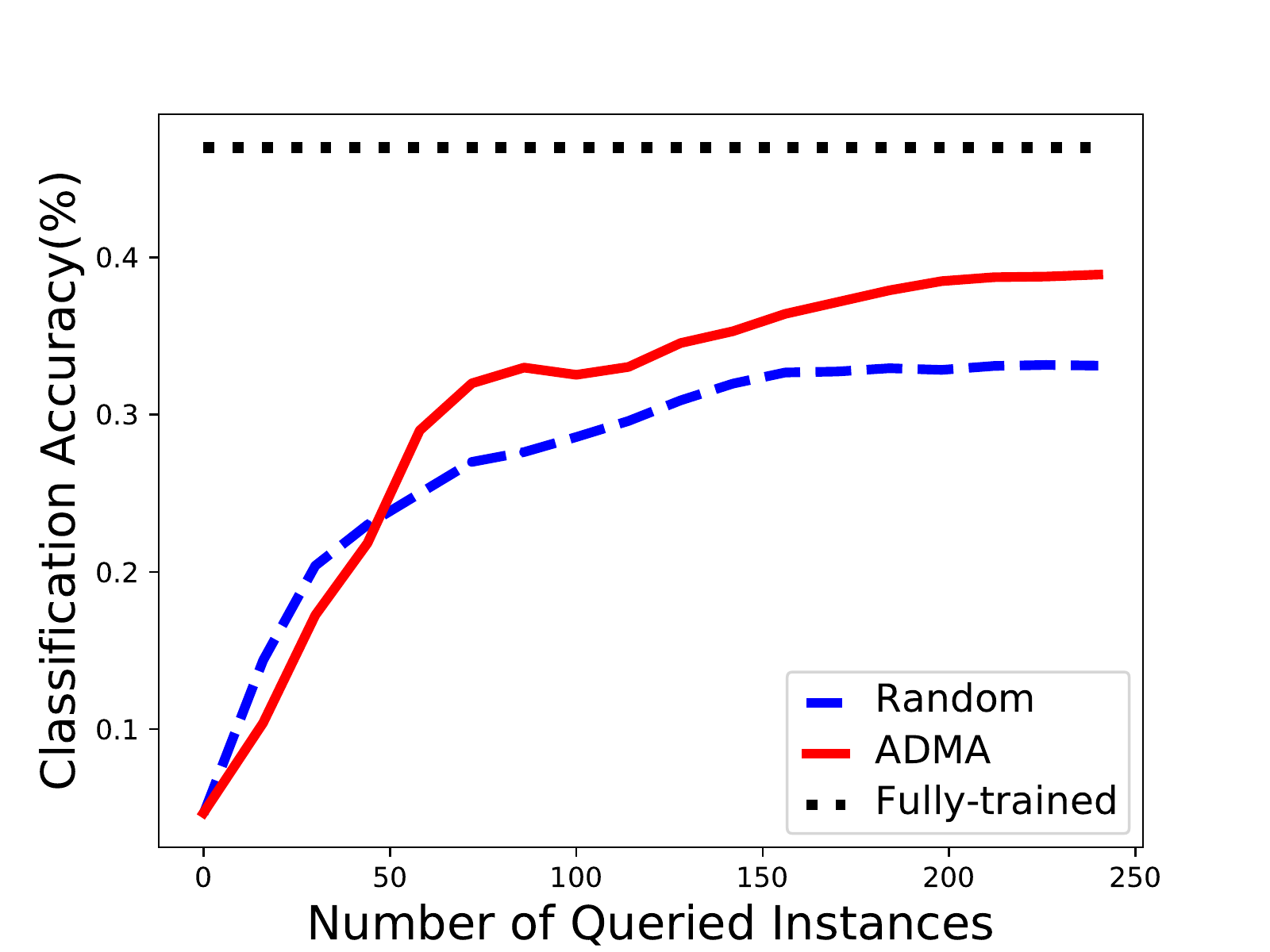}\\
			\centering{(a) AlexNet + PASCAL VOC2012}
		\end{minipage}
		\begin{minipage}{0.32\linewidth}
			\includegraphics[width=\textwidth]{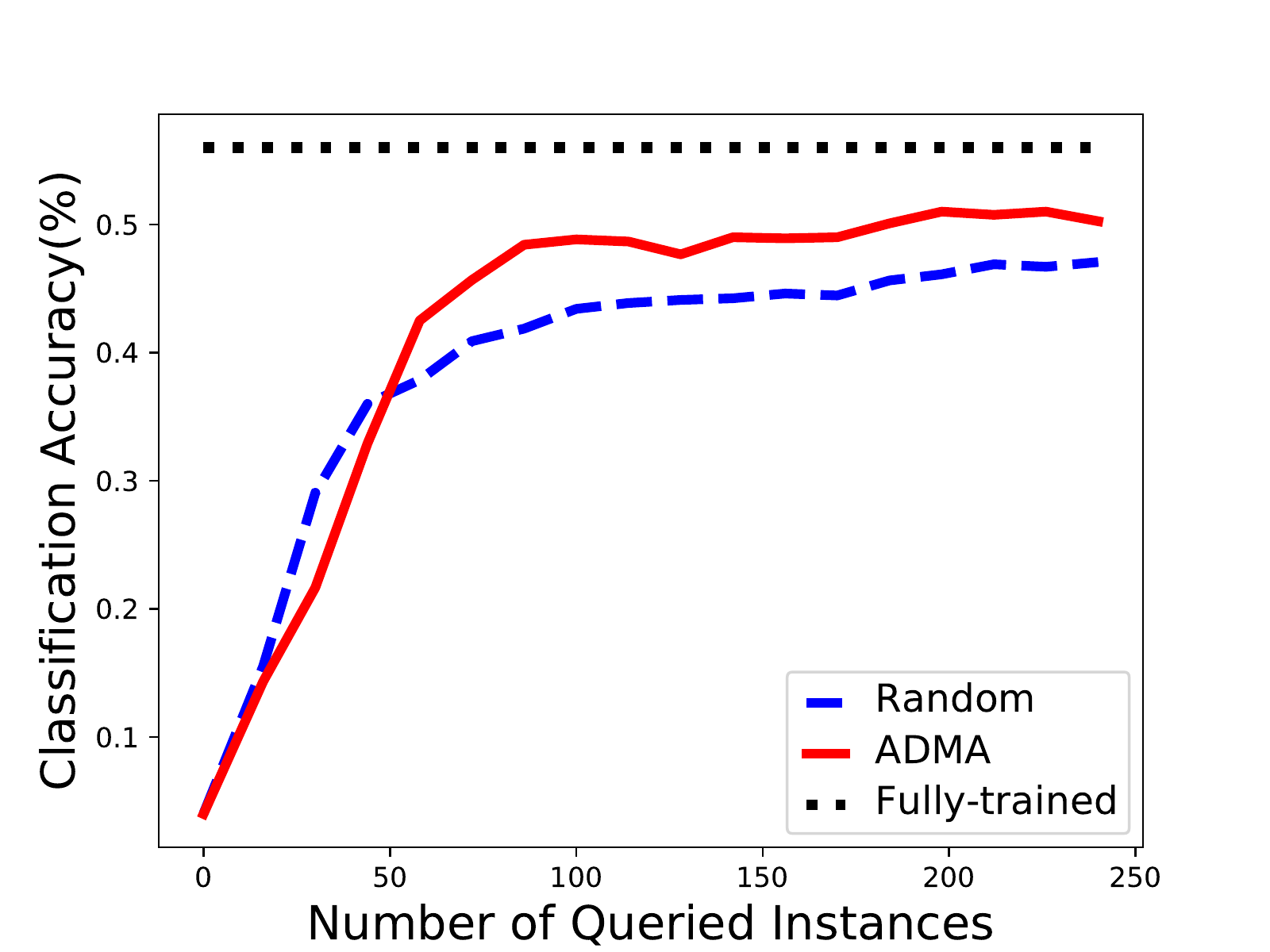}\\
			\centering{(b) VGG-16 + PASCAL VOC2012}
		\end{minipage}      
		\begin{minipage}{0.32\linewidth}
			\includegraphics[width=\textwidth]{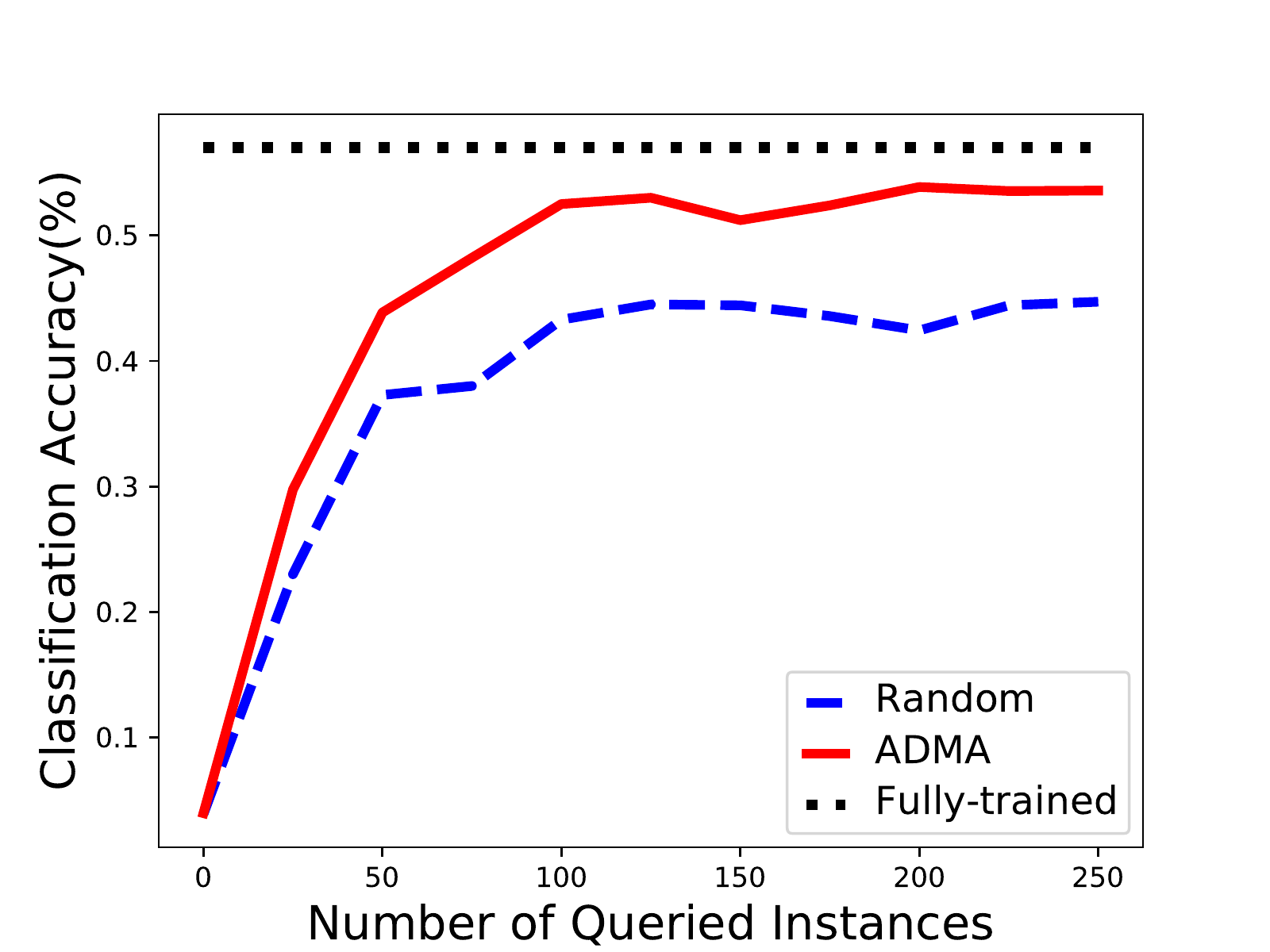}\\
			\centering{(c) ResNet-18 + PASCAL VOC2012}
		\end{minipage}
		
		\begin{minipage}{0.32\linewidth}
			\includegraphics[width=\textwidth]{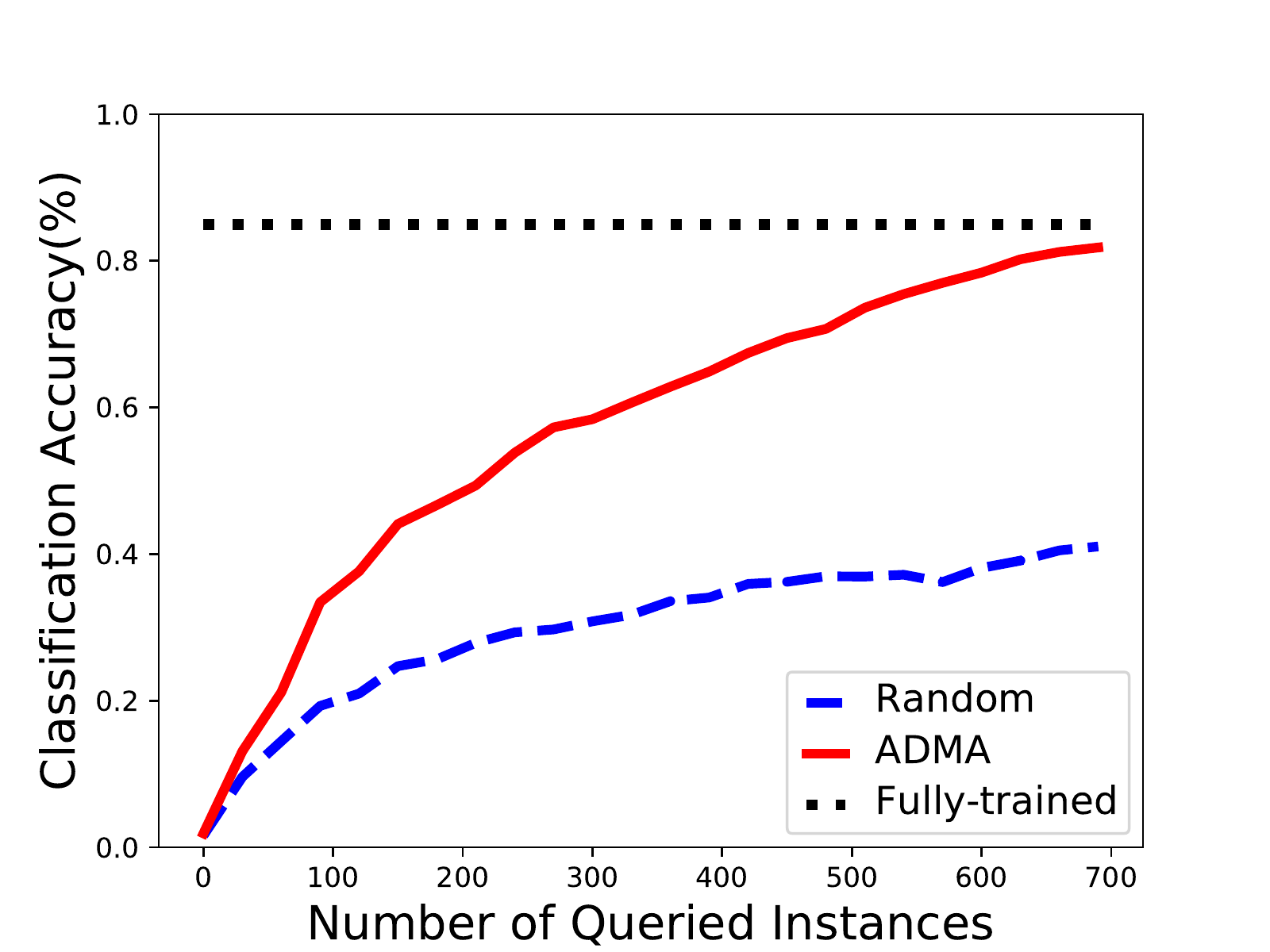}\\
			\centering{(d) AlexNet + Indoor}
		\end{minipage}
		\begin{minipage}{0.32\linewidth}
			\includegraphics[width=\textwidth]{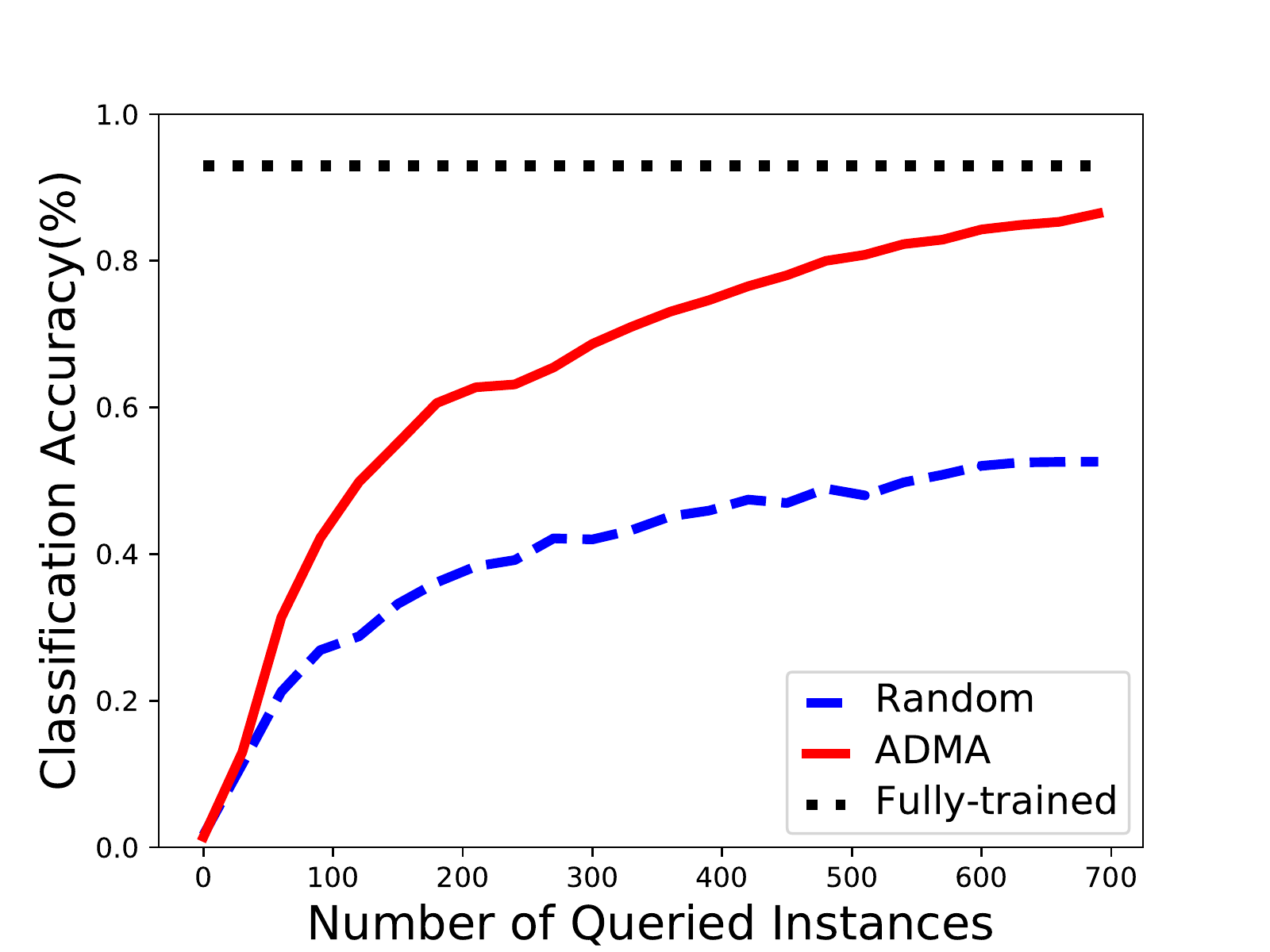}\\
			\centering{(e) VGG-16 + Indoor}
		\end{minipage}      
		\begin{minipage}{0.32\linewidth}
			\includegraphics[width=\textwidth]{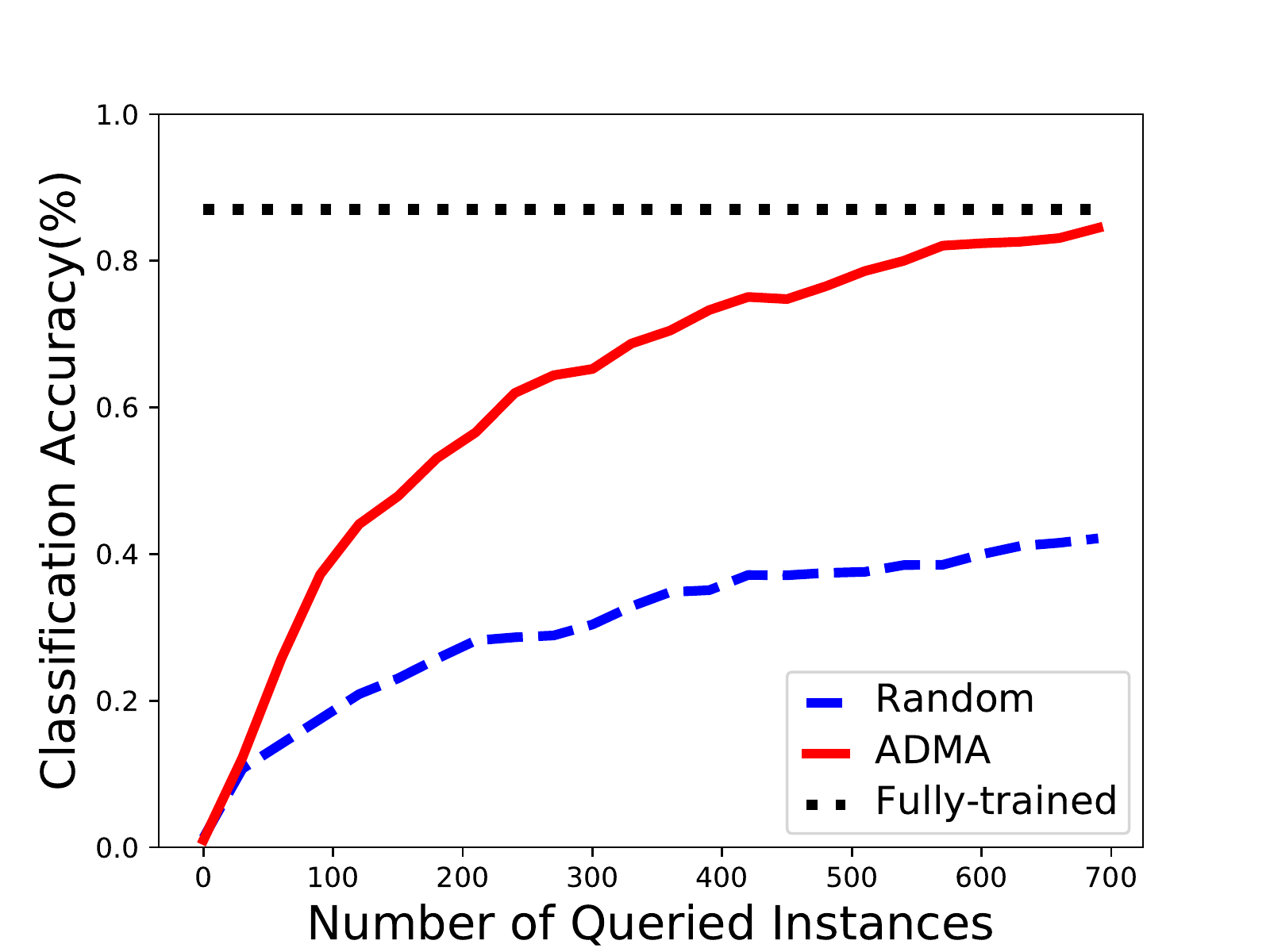}\\
			\centering{(f) ResNet-18 + Indoor}
		\end{minipage}
		\caption{The comparison results on multi-class datasets with different pre-trained models.}\label{fig:multiclass}       
	\end{center}
\end{figure*}

\section{Experiments}
In the experiments, we study with the following three pre-trained models:
\begin{itemize}
	\item AlexNet \cite{18} is the winning solution of ImageNet Challenge 2012 which has 15.4\% top-5 error on ILSVRC dataset. The network consists of 5 conv layers, max-pooling layers, dropout layers, and 3 fully connected layers. 
	\item VGG \cite{19} is the model designed for ImageNet Challenge 2014, which has 7.3\% top-5 error on ILSVRC dataset. This network is characterized by its simplicity, using only $3\times 3$ convolutional layers stacked on top of each other with increasing depth. Max pooling is used to reduce the volume size. We employ VGG-16 in Pytorch for implementation.
	\item ResNet \cite{20} is the winning solution of ImageNet Challenge 2015 which has 3.6\% top-5 error on ILSVRC dataset. Its architecture contains both plain network and residual network. We employ ResNet-18 in Pytorch for implementation.
\end{itemize}

When applying the pre-trained models to new tasks, the number of nodes in the last layer is changed to fit the number of classes. For all the pre-trained models, the second last layer is specified as layer $B$, while the fourth layer from the end is specified as layer $A$ in our approach.

We also introduce the datasets involved. All the above introduced models are pre-trained on the ImageNet ILSVRC2012 dataset. In Section 4.1, we will perform experiments on two multi-class datasets and two binary classification datasets. PASCAL VOC2012 \cite{21} is an well known image dataset of visual objects in realistic scenes. It consists of 17,125 images from 20 classes. Indoor \cite{22} is a dataset with a total of 15,620 images from 67 indoor categories. DOGvsCAT is a dataset from kaggle for classifying 25,000 images between dog and cat \cite{23}. INRIA Person Dataset is a dataset with a total of 1832 images to classify person from it\cite{24}. All the images are resized to $224\times 224$ in order to fit the input of pre-trained models.

\subsection{Performance comparison}


\begin{table*}[tp]
	\centering
	\begin{threeparttable}
		\caption{AUC results on Pascal VOC2012.}
		\label{table:auc1}
		\begin{tabular}{cccccccc}
			\toprule
			\multirow{2}{*}{Models} &\multirow{2}{*}{Algorithms}
			&\multicolumn{6}{c}{Number of queried instances}\cr
			\cmidrule(lr){3-8} 
			& &20 &40 &60 &80 &100 &120\cr
			\midrule
			\multirow{2}{*}{AlexNet} &ADMA &0.676($\pm0.003$)&0.756($\pm0.025$)&{\bfseries 0.787($\pm0.025$)}&{\bfseries 0.808($\pm0.023$)} &{\bfseries 0.823($\pm0.024$)}&{\bfseries 0.823($\pm0.024$)}\cr
			&RANDOM
			&{\bfseries 0.725($\pm0.012$)}& {\bfseries 0.767($\pm0.018$)} & 0.778($\pm0.010$)&0.795($\pm0.006$)&0.806($\pm0.004$)&0.807($\pm0.004$)\cr
			\hline
			\multirow{2}{*}{VGG-16} &ADMA
			&0.748($\pm0.018$)&{\bfseries 0.863($\pm0.002$)}&{\bfseries 0.886($\pm0.000$)}&{\bfseries 0.891($\pm0.002$)}&{\bfseries 0.896($\pm0.002$)}&{\bfseries 0.897($\pm0.002$)}\cr
			&RANDOM
			&{\bfseries 0.748($\pm0.013$)}&0.818($\pm0.010$)&0.855($\pm0.019$)&0.870($\pm0.003$)&0.879($\pm0.003$)&0.880($\pm0.004$)\cr
			\hline
			\multirow{2}{*}{ResNet-18} &ADMA
			&{\bfseries 0.820($\pm0.034$)}&0.876($\pm0.005$)&0.894($\pm0.004$)&{\bfseries 0.907($\pm0.004$)}&{\bfseries 0.907($\pm0.001$)}&{\bfseries 0.909($\pm0.000$)}\cr
			&RANDOM
			&0.805($\pm0.006$)&{\bfseries 0.886($\pm0.001$)}&{\bfseries 0.895($\pm0.000$)}&0.900($\pm0.001$)&0.903($\pm0.001$)&0.907($\pm0.002$)\cr
			\bottomrule
		\end{tabular}
	\end{threeparttable}
\end{table*}
\begin{table*}[tp]
	
	\centering
	\begin{threeparttable}
		\caption{AUC results on Indoor.}
		\label{table:auc2}
		\begin{tabular}{ccccccccc}
			\toprule
			\multirow{2}{*}{Models} &\multirow{2}{*}{Algorithms}
			&\multicolumn{6}{c}{Number of queried instances}\cr
			\cmidrule(lr){3-8}
			& &100 &200 &300 &400 &500 &600 \cr
			\midrule
			\multirow{2}{*}{AlexNet} &ADMA &{\bfseries 0.858($\pm0.014$)}&{\bfseries 0.924($\pm0.013$)}&{\bfseries 0.945($\pm0.026$)}&{\bfseries 0.958($\pm0.005$)}&{\bfseries 0.964($\pm0.003$)}&{\bfseries0.967($\pm0.005$)}\cr
			&RANDOM
			&0.716($\pm0.013$)&0.781($\pm0.010$)&0.820($\pm0.020$)&0.847($\pm0.005$)&0.863($\pm0.003$)&0.888($\pm0.005$)\cr
			\hline
			\multirow{2}{*}{VGG-16} &ADMA
			&{\bfseries 0.906($\pm0.001$)}&{\bfseries 0.960($\pm0.006$)}&{\bfseries 0.971($\pm0.003$)}&{\bfseries 0.976($\pm0.003$)}&{\bfseries 0.980($\pm0.001$)}&{\bfseries0.983($\pm0.009$)}\cr
			&RANDOM
			&0.766($\pm0.007$)&0.839($\pm0.006$)&0.871($\pm0.009$)&0.908($\pm0.001$)&0.922($\pm0.003$)&0.929($\pm0.003$)\cr
			\hline
			\multirow{2}{*}{ResNet-18} &ADMA
			&{\bfseries 0.834($\pm0.016$)}&{\bfseries 0.938($\pm0.002$)}&{\bfseries 0.961($\pm0.002$)}&{\bfseries 0.9731($\pm0.001$)}&{\bfseries 0.977($\pm0.001$)}&{\bfseries 0.981($\pm0.002$)}\cr
			&RANDOM
			&0.685($\pm0.013$)&0.781($\pm0.004$)&0.826($\pm0.002$)&0.852($\pm0.004$)&0.866($\pm0.007$)&0.877($\pm0.007$)\cr
			\bottomrule
		\end{tabular}
	\end{threeparttable}
\end{table*}

\begin{figure*}[!h]
	\begin{center}
		\begin{minipage}{0.32\linewidth}
			\includegraphics[width=\textwidth]{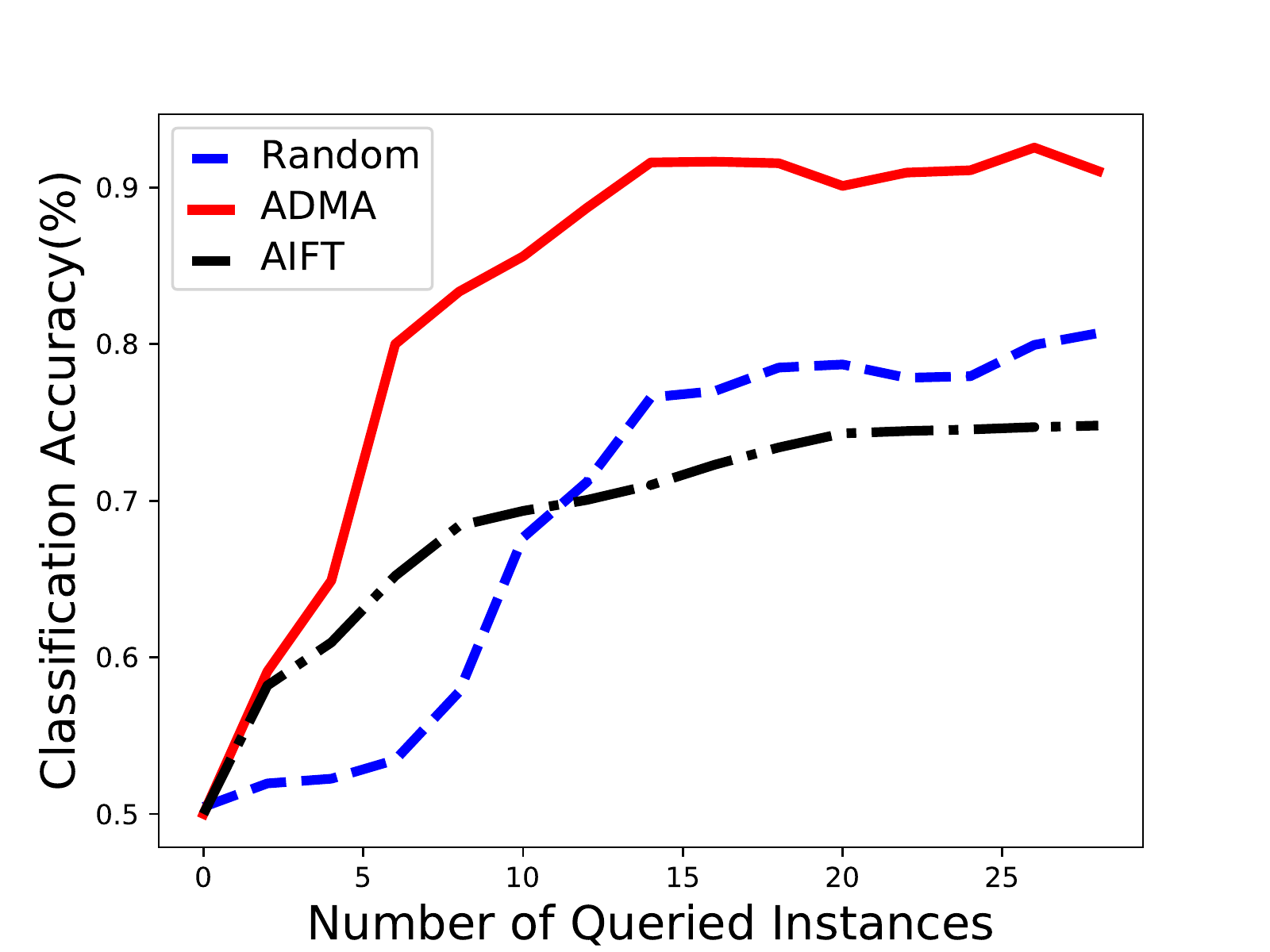}\\
			\centering{(a) AlexNet + DOG vs CAT}
		\end{minipage}
		\begin{minipage}{0.32\linewidth}
			\includegraphics[width=\textwidth]{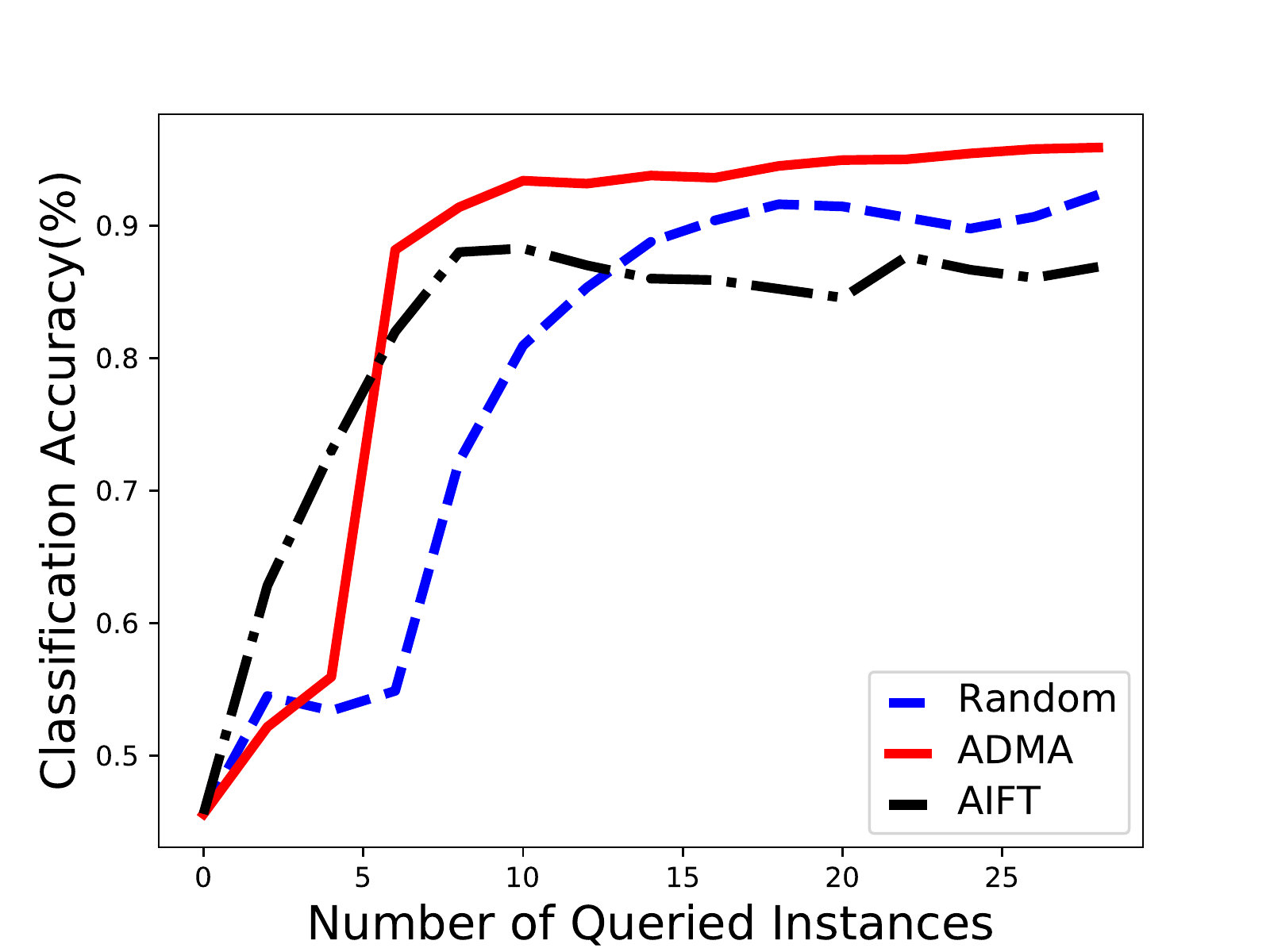}\\
			\centering{(b) VGG-16 + DOG vs CAT}
		\end{minipage}      
		\begin{minipage}{0.32\linewidth}
			\includegraphics[width=\textwidth]{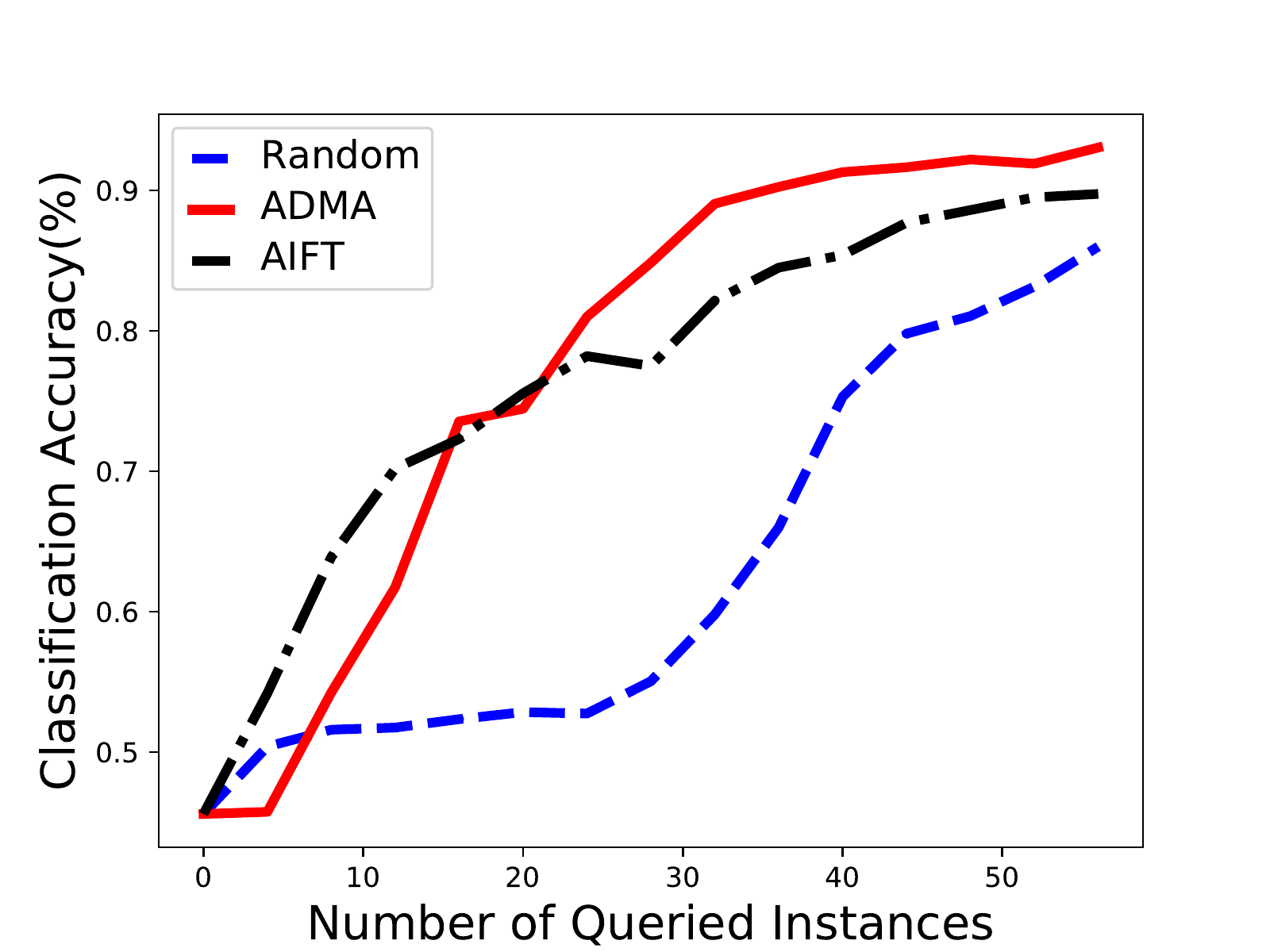}\\
			\centering{(c) ResNet-18 + DOG vs CAT}
		\end{minipage}
		
		\begin{minipage}{0.32\linewidth}
			\includegraphics[width=\textwidth]{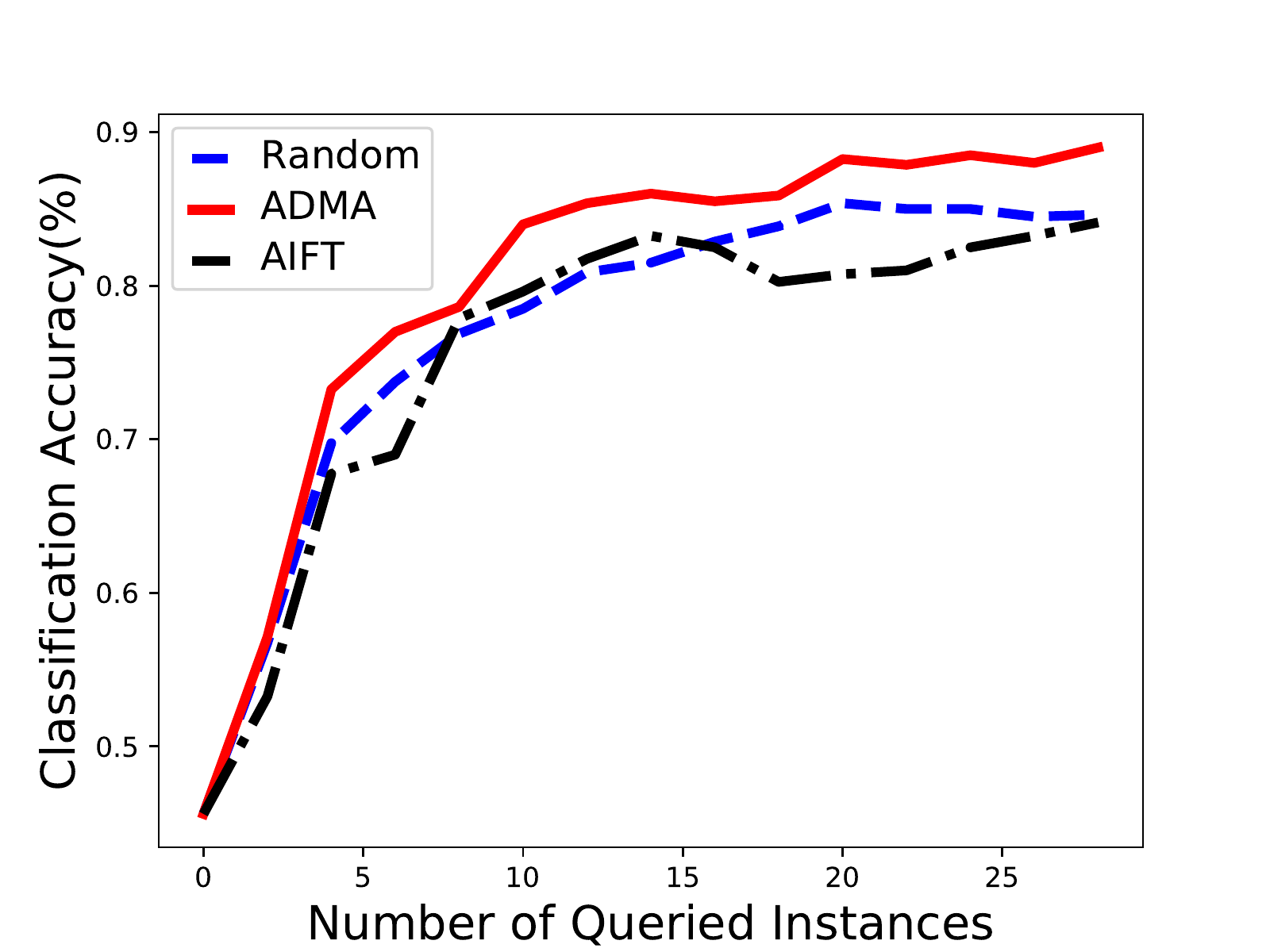}\\
			\centering{(d) AlexNet + INRIA Person Dataset}
		\end{minipage}
		\begin{minipage}{0.32\linewidth}
			\includegraphics[width=\textwidth]{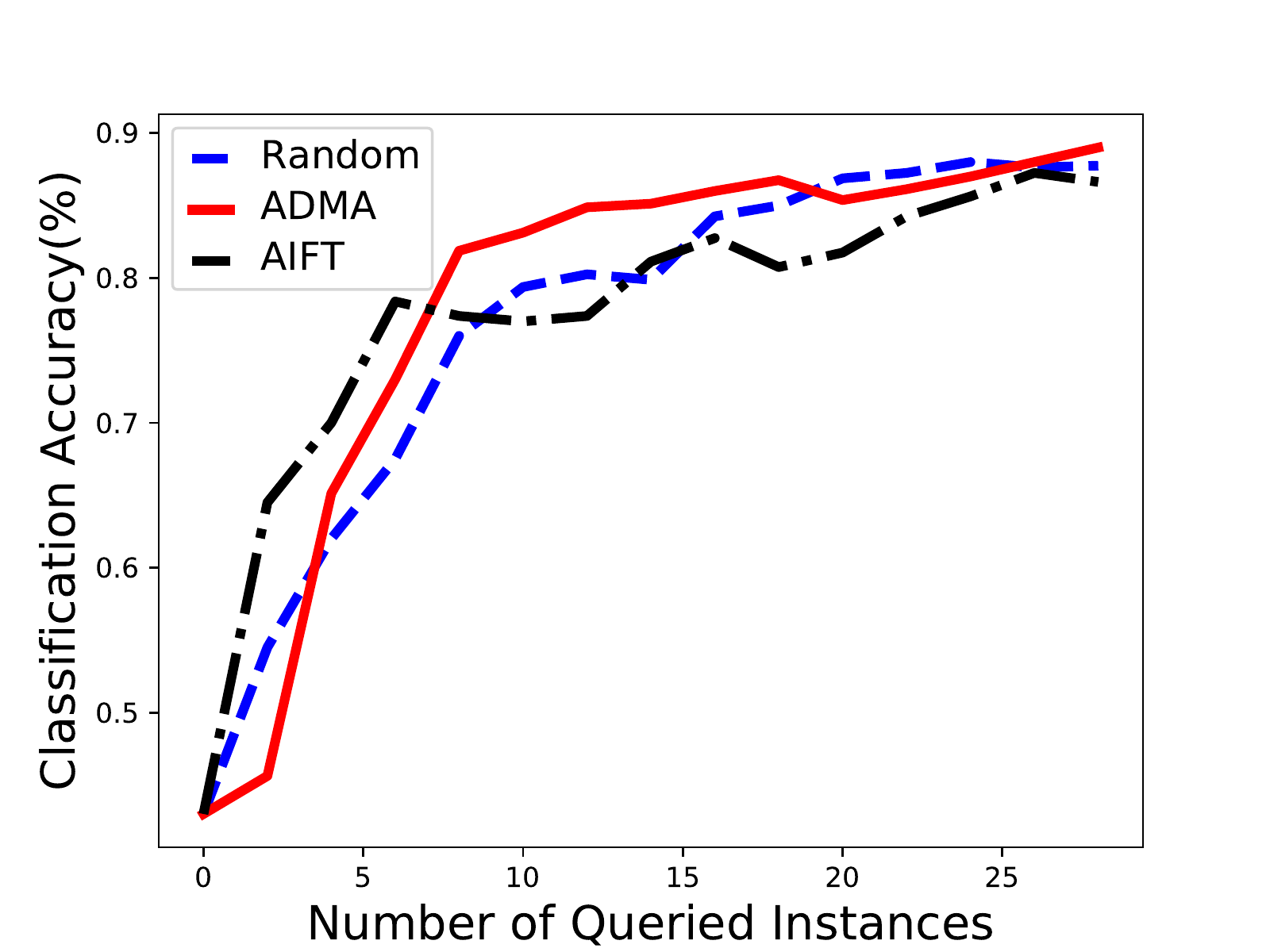}\\
			\centering{(e) VGG-16 + INRIA Person Dataset}
		\end{minipage}      
		\begin{minipage}{0.32\linewidth}
			\includegraphics[width=\textwidth]{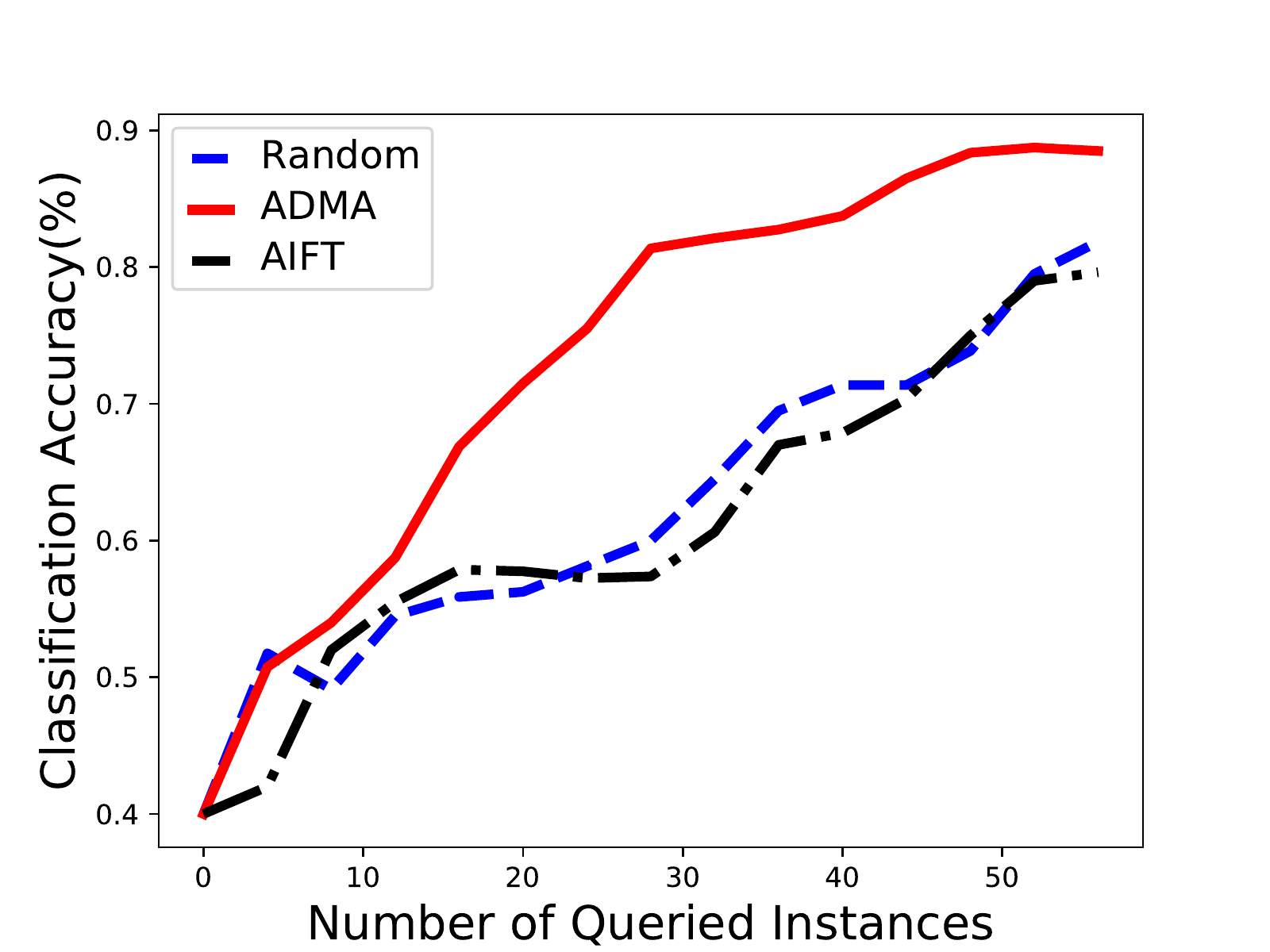}\\
			\centering{(f) ResNet-18 + INRIA Person Dataset}
		\end{minipage}
		\caption{The comparison results on binary-class datasets with different pre-trained models.}\label{fig:binary}       
	\end{center}
\end{figure*}

We perform the comparison in both multi-class and binary classification tasks. Note that there is no general approach applicable to our setting, we compare the proposed approach with the Random method, which randomly select instances to query their labels. The method AIFT proposed in \cite{13} was originally designed for binary classification of medical images, and thus is not compared in the multi-class cases. Instead, the performance of the fully re-trained model is provided for reference. This model use exactly the same architecture of our method but is re-trained with all data from the target task as labeled.

We freeze the first two layers for AlexNet, the first seven layers for VGG-16, and the first eight layers for ResNet-18, and fine tune the other layers. For multi-class problems, the batch size is set to 2 and 10 for VOC and Indoor, respectively. Instead, we query one label at each time for binary tasks, because they are relatively simpler. For VOC and INRIA datasets, we follow the original partition to separate training and test set. For the other two datasets, we select 70\% examples as the unlabeled pool for active selection, and the rest 30\% as the test set to validate the classification performance.


For multi-class tasks, the accuracy curves are plotted in Figure \ref{fig:multiclass}. It can be observed that with different pre-trained models and different datasets, our algorithm consistently outperforms the random sampling method. The results validate the effectiveness of our approach. Especially, it is surprise to observe that by querying less than 5\% examples of the dataset, our approach ADMA can achieve comparable performance with the fully trained model using all examples.

We also report the AUC results in Tables \ref{table:auc1} and \ref{table:auc2}. AUC is commonly used to evaluate the performance of multi-class classification. The results are generally consistent with that in Figure \ref{fig:multiclass}.

For binary classification tasks, the accuracy curves are plotted in Figure \ref{fig:binary}. Because the task is relatively easier than that of multi-class learning, the performance curves increase more fast. The proposed ADMA approach can outperforms the other two methods in most cases. The performance of AIFT is mixed. It achieves decent performance with ResNet-18 on DOGvsCAT, and with VGG-16 on INRIA dataset, but loses its edge on the for the other cases.

\begin{figure}[!tp]
	\includegraphics[width=0.38\textwidth]{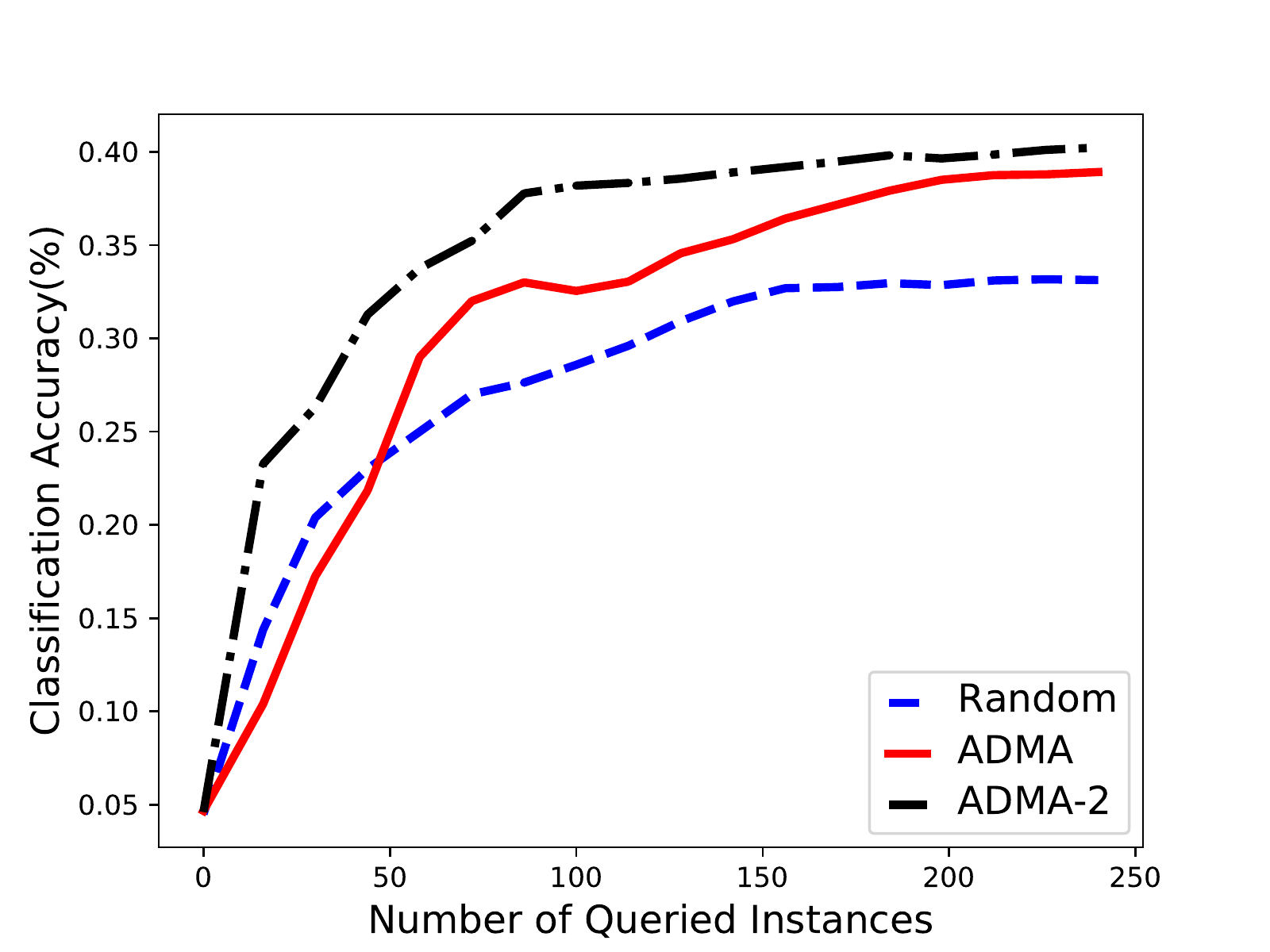}\\
	\caption{Comparison results of multiple transform patterns.}\label{fig:multilayer}
\end{figure}

\begin{figure}[!b]
	\begin{center}
		\begin{minipage}{0.46\linewidth}
			\includegraphics[width=\textwidth]{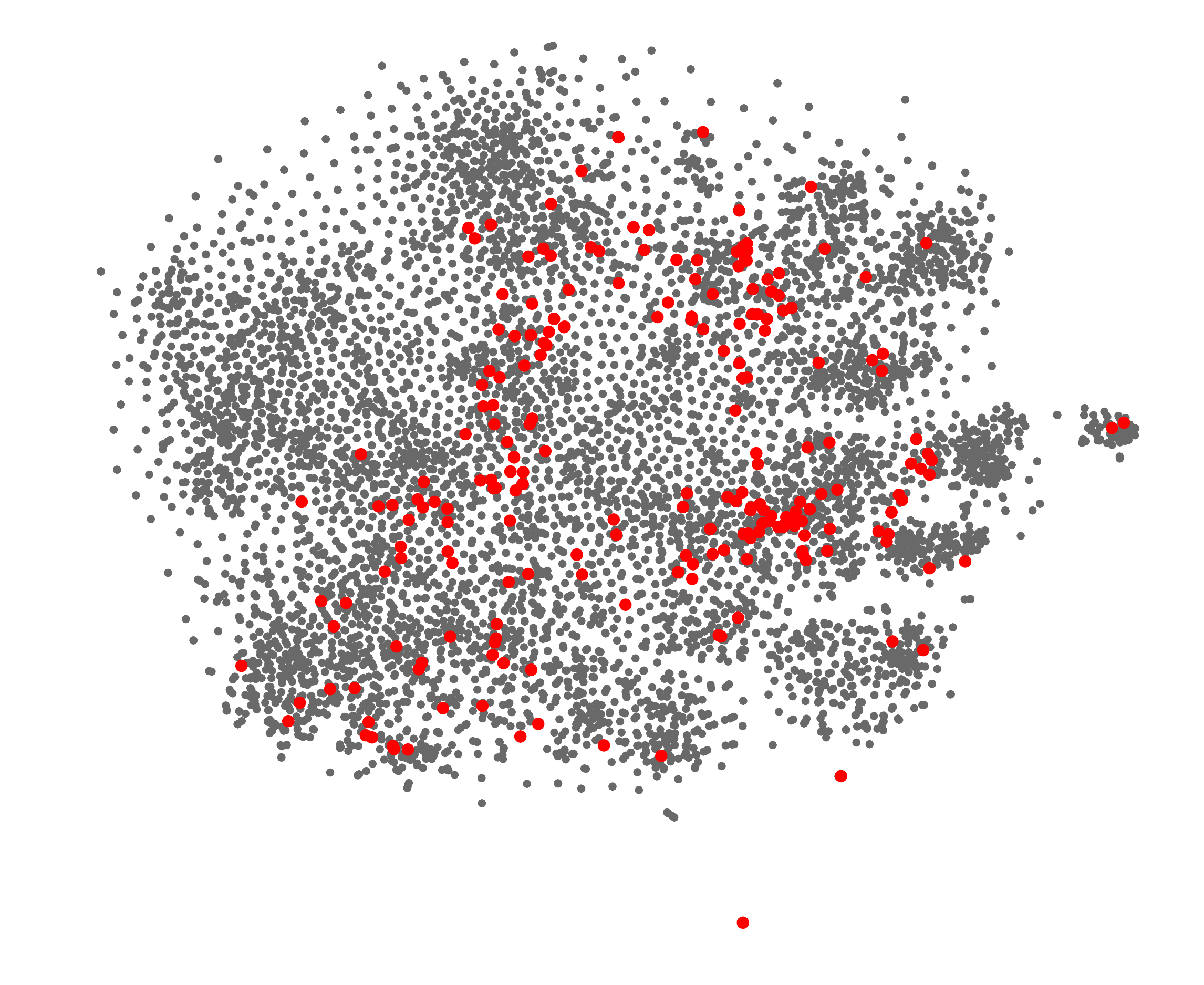}\\
			\centering{(a) distribution of queries by ADMA\\$ $}
		\end{minipage}$\quad$
		\begin{minipage}{0.46\linewidth}
			\includegraphics[width=\textwidth]{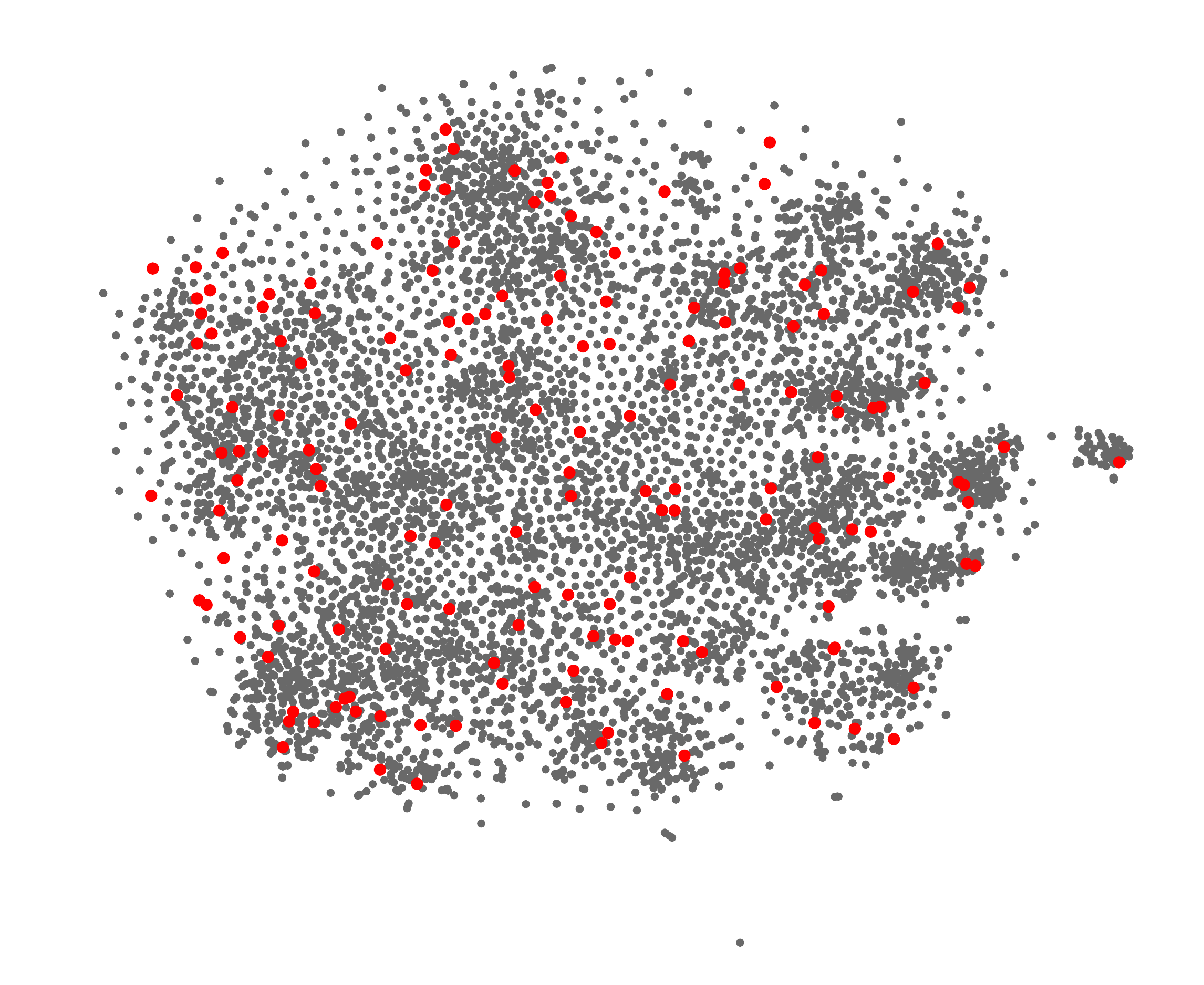}\\
			\centering{(b) distribution of queries by Random\\$ $}
		\end{minipage}  
		
		\begin{minipage}{0.46\linewidth}
			\includegraphics[width=\textwidth]{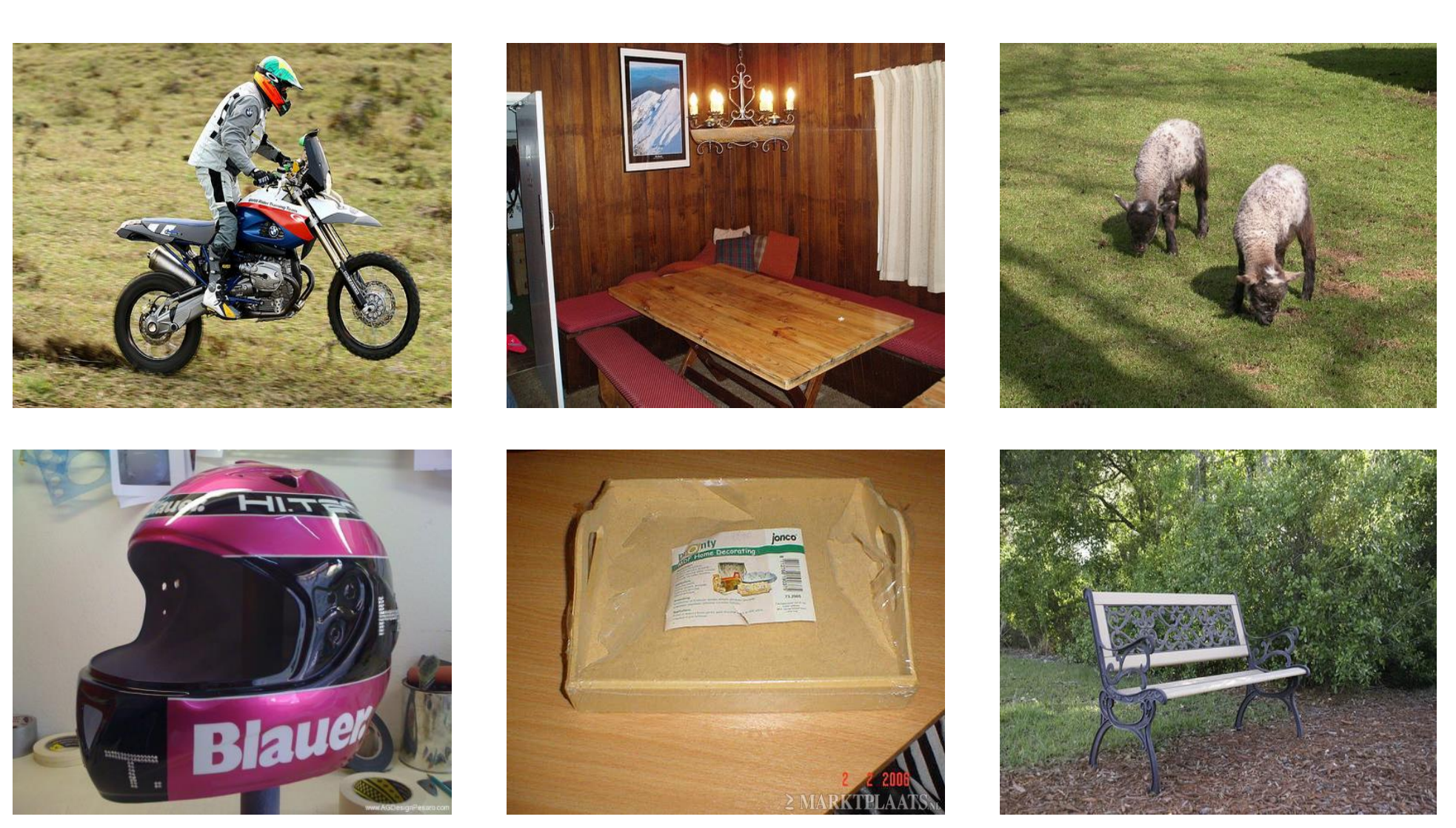}\\
			\centering{(c) typical examples of queried images}
		\end{minipage} $\quad$
		\begin{minipage}{0.46\linewidth}
			\includegraphics[width=\textwidth]{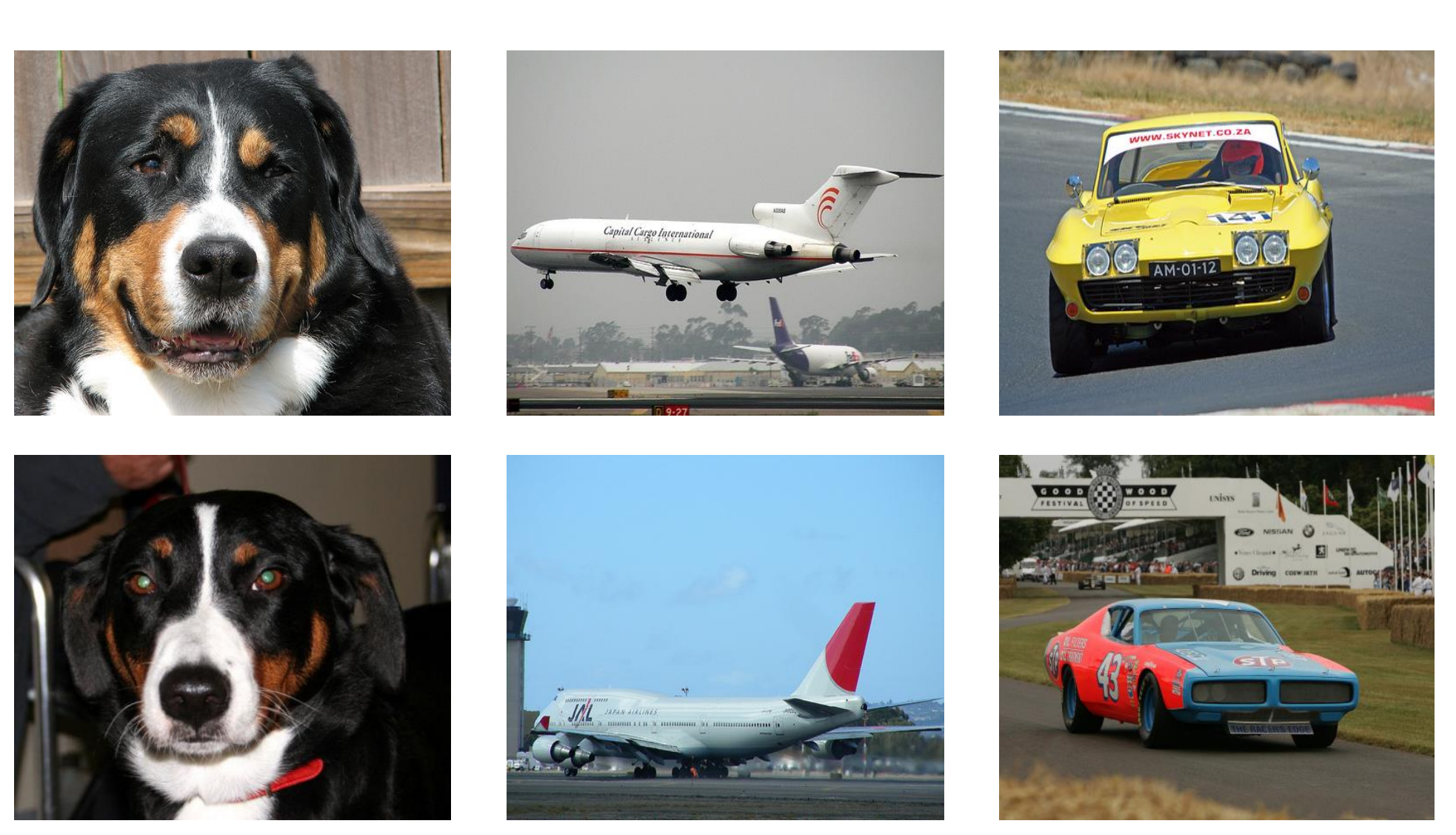}\\
			\centering{(d) typical examples of not-queried images.}
		\end{minipage} 
		\caption{Visualization of the queried images.}\label{fig:example} 
	\end{center}      
\end{figure}

\subsection{Study on the distinctiveness}
The proposed criterion \emph{distinctiveness} is calculated based on the feature transformation pattern from layer $A$ to layer $B$. In this subsection, we further examine whether the performance can be improved by exploiting multiple patterns. We take AlexNet on PASCAL VOC2012 dataset as an example to perform the experiment. Specifically, we set two candidate starting layers, i.e. the fourth layer from the end as $A_1$ and the fifth layer from the end as $A_2$. Then two transformation patterns $A_1\rightarrow B$ and $A_2\rightarrow B$ are calculated, and the variance between them is used to estimate the \emph{distinctiveness}. 

The comparison results are plotted in Figure \ref{fig:multilayer}. By exploiting multiple transform patterns between different layer pairs, ADMA-2 can further improve the performance. While we only examine the case with two patterns, it can be expected to achieve further improvements with the ensemble of more layer pairs.

In addition, we try to examine whether the query results well match our motivation that distinctive instances should be preferred. We visualize the data points via t-SNE \cite{25} based on the representation of layer $B$. Figure \ref{fig:example} (a) and (b) show the queried examples by the proposed ADMA approach and random sampling, respectively. Obviously, the queried images of ADMA have a biased distribution while the queries of random sampling follow a uniform distribution. We further show some typical images among the queried and not-queried examples of our approach in Figure \ref{fig:example} (c) and (d), respectively. For each pair, the first row is the image queried/not-queried image in the target task, while the second row is the closest center image in the source task. For the queried group, the images in the source and target task have distinctive difference, while images in the not-queried group look similar. This observation is consistent with our motivation on \emph{distinctiveness} that the images capture the unique property of the target task should be better queried.
%

\begin{figure}[!tb]
	\includegraphics[width=0.38\textwidth]{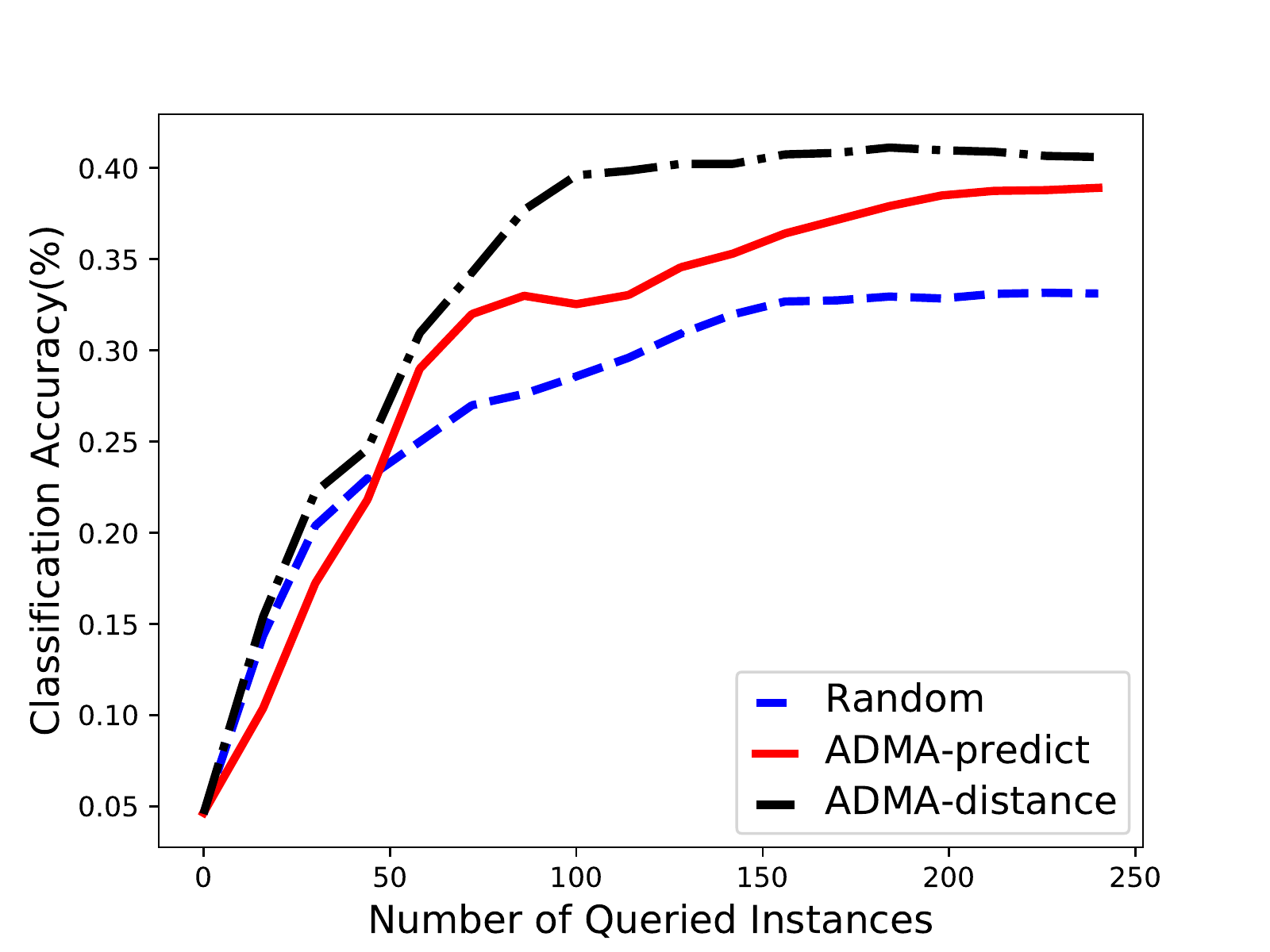}\\
	\caption{Comparison results of approximation weights.}\label{fig:weight}
\end{figure}

\subsection{Study on the approximation weights}
Lastly, we examine the other possible solutions for the weighs $\alpha_k(\bx)$ in Eq. \ref{eqn:Shx} when approximating the transformation pattern of $\bx$ with weighted combination of source patterns. In fact, the weight $\alpha_k(\bx)$ is estimating how likely $\bx$ is from the $k$-th class if we assume it is an example in the source task. In all the above experiments, we calculate $\alpha_k(\bx)$ as Eq. \ref{eqn:alpha}, i.e., use the prediction probability on the $k$-th class as the weight corresponding to the $k$-th center. We denote the method with this implementation as ADMA-predict. Here we provide another solution, which calculates the weight $\alpha_k(\bx)$ by the reciprocal $L$-2 distance between $\bx$ and $\bm c_k$ based on the representation of layer $A$. We denote by ADMA-distance. Again we perform the experiment with AlexNet on PASCAL VOC2012, and plot the results in Figure \ref{fig:weight}. It can be observed that ADMA-distance outperforms ADMA-predict consistently. This is probably because that the pre-trained model may be not reliable if it is directly employed to predict an example from a different task. This is also the reason we need to perform model adaptation across tasks.

\section{Conclusion}
In this paper, we propose an active model adaptation approach for cost-effective training of deep convolutional neural networks. Instead of training from scratch, a pre-trained model can be effectively adapted to a new target task by fine tuning with a few actively queried examples, significantly reducing the cost of designing the network architecture and labeling a large training set. To select the most useful instances for label querying, a novel active selection criterion is proposed, which dynamically balances between distinctiveness and uncertainty. The distinctiveness measures the potential contribution of an instance on improving the network for better feature representation, while uncertainty measures the capability on improving the classifier of the last layer. Experiments are performed on different datasetes with different pre-trained models. The results show that the proposed approach can achieve effective deep network training with significantly lower cost. In the future, we plan to apply the approach on more datasets and more pre-trained models. Also, the distinctiveness criterion will be further studied by considering the feature transformation patterns among different layers.

\section*{Acknowledgment}
This research was partially supported by JiangsuSF (BK20150754), NSFC (61503182, 61732006) and China Postdoctoral Science Foundation.

\bibliographystyle{ACM-Reference-Format}
\bibliography{refer}

\end{document}